\documentclass{article}

\usepackage{microtype}
\usepackage{booktabs} % for professional tables
\usepackage{enumitem}
\usepackage{caption}
\usepackage{subcaption}
\usepackage{url}   
\usepackage{amsfonts}       % blackboard math symbols
\usepackage{nicefrac}       % compact symbols for 1/2, etc.
\usepackage{xcolor} 
\usepackage{hyperref}

\usepackage[accepted]{icml2024}
\usepackage[left=1.2in, right=1.2in, top=1.in, bottom=1.2in]{geometry}

\usepackage{amsmath}
\usepackage{amssymb}
\usepackage{mathtools}
\usepackage{amsthm}
\usepackage{amsmath,amsfonts,bm}

\def\eqref#1{equation~\ref{#1}}
\def\1{\bm{1}}

\DeclareMathAlphabet{\mathsfit}{\encodingdefault}{\sfdefault}{m}{sl}
\SetMathAlphabet{\mathsfit}{bold}{\encodingdefault}{\sfdefault}{bx}{n}

\newcommand{\E}{\mathbb{E}}

\usepackage[capitalize,noabbrev]{cleveref}

\theoremstyle{plain}
\newtheorem{theorem}{Theorem}[section]

\theoremstyle{definition}

\theoremstyle{remark}

\newtheorem*{theorem*}{Theorem}

\usepackage[textsize=tiny]{todonotes}

\icmltitlerunning{Insignificance of SGD Noise in Online Learning}

\begin{document}

\icmltitle{Beyond Implicit Bias:
       The Insignificance of SGD Noise \newline in Online Learning}

% It is OKAY to include author information, even for blind
% submissions: the style file will automatically remove it for you
% unless you've provided the [accepted] option to the icml2024
% package.

% List of affiliations: The first argument should be a (short)
% identifier you will use later to specify author affiliations
% Academic affiliations should list Department, University, City, Region, Country
% Industry affiliations should list Company, City, Region, Country

% You can specify symbols, otherwise they are numbered in order.
% Ideally, you should not use this facility. Affiliations will be numbered
% in order of appearance and this is the preferred way.
\icmlsetsymbol{equal}{*}

\begin{icmlauthorlist}
\icmlauthor{Nikhil Vyas}{equal,yyy}
\icmlauthor{Depen Morwani}{equal,yyy}
\icmlauthor{Rosie Zhao}{equal,yyy}
\icmlauthor{Gal Kaplun}{equal,yyy}
\icmlauthor{Sham Kakade}{yyy,zzz}
\icmlauthor{Boaz Barak}{yyy}
\end{icmlauthorlist}

\icmlaffiliation{yyy}{SEAS, Harvard University}
\icmlaffiliation{zzz}{Kempner Institute, Harvard University}

\icmlcorrespondingauthor{Nikhil Vyas}{nikhil@g.harvard.edu}
\icmlcorrespondingauthor{Depen Morwani}{dmorwani@g.harvard.edu}

% You may provide any keywords that you
% find helpful for describing your paper; these are used to populate
% the "keywords" metadata in the PDF but will not be shown in the document
\icmlkeywords{online learning, SGD noise, implicit bias}

\vskip 0.3in

% this must go after the closing bracket ] following \twocolumn[ ...

% This command actually creates the footnote in the first column
% listing the affiliations and the copyright notice.
% The command takes one argument, which is text to display at the start of the footnote.
% The \icmlEqualContribution command is standard text for equal contribution.
% Remove it (just {}) if you do not need this facility.

%\printAffiliationsAndNotice{}  % leave blank if no need to mention equal contribution
\printAffiliationsAndNotice{\icmlEqualContribution} % otherwise use the standard text.

\begin{abstract}
The success of SGD in deep learning has been ascribed by prior works to the \emph{implicit bias} induced by finite batch sizes (``SGD noise''). While prior works focused on \emph{offline learning} (i.e., multiple-epoch training), we study the impact of SGD noise on \emph{online} (i.e., single epoch) learning. Through an extensive empirical analysis of image and language data, we demonstrate that small batch sizes do \emph{not} confer any implicit bias advantages in online learning. In contrast to offline learning, the benefits of SGD noise in online learning are strictly computational, facilitating more cost-effective gradient steps. This suggests that SGD in the online regime can be construed as taking noisy steps along the ``golden path'' of the noiseless \emph{gradient descent} algorithm. We study this hypothesis and provide supporting evidence in loss and function space. Our findings challenge the prevailing understanding of SGD and offer novel insights into its role in online learning.
\end{abstract}

\section{Introduction}\label{sec:intro}
In the field of optimization theory, the selection of hyperparameters, such as learning rate and batch size, plays a significant role in determining the \emph{optimization efficiency}, which refers to the computational resources required to minimize the loss function to a predetermined level. In strongly \emph{convex problems}, altering these hyperparameters does not affect the final solution since all local minima are global. 
Hence, the final model only depends on the \emph{explicit biases} of architecture and objective function (including any explicit regularizers).
In contrast, Deep Learning is \emph{non-convex}, which means that the choices of algorithm and hyperparameters can impact not only optimization efficiency but also introduce an \emph{implicit bias}, i.e., change the regions of the search space explored by the optimization algorithm, consequently impacting the final learned model.

The implicit bias induced by the algorithm and hyperparameter choices can significantly affect the quality of the learned model, including generalization, robustness to distribution shifts, downstream performance, and more. 
Hence, the implicit bias of \emph{stochastic gradient descent (SGD)} has garnered considerable attention within the research community \citep{Jastrzebski,catapult, damian2021, control-bs-lr, pmlr-v162-nacson22a, haochen21a,andriushchenko2022sgd}. In particular, this implicit bias emerges due to SGD using finite batch sizes resulting in a \emph{noisy approximation} of the population gradient.

\begin{figure*}[!htb]
\hspace*{-1.2cm} 
    \centering
    \includegraphics[width=5.2in]{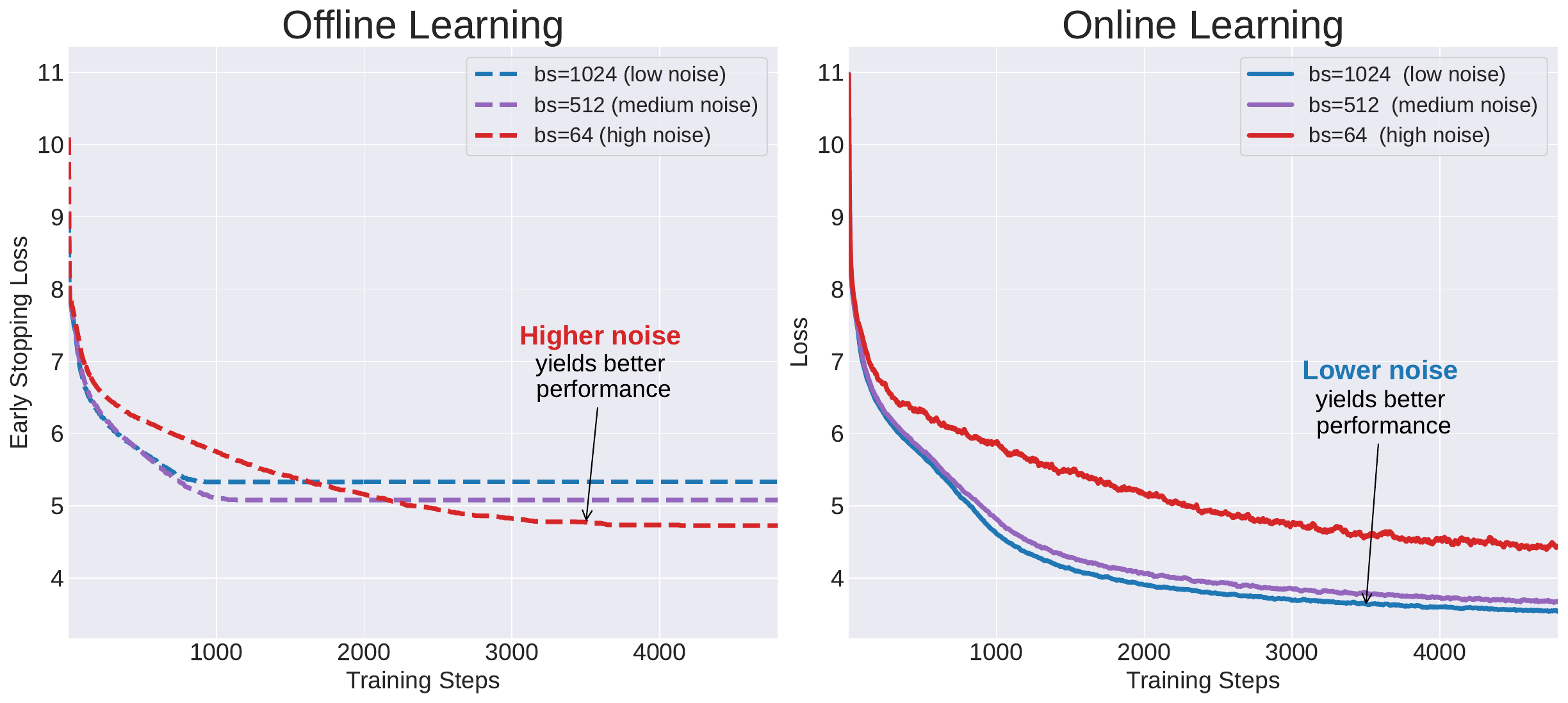}
    \caption{Experiment on offline (\textbf{left}) and online (\textbf{right}) learning on the C4 dataset across various  batch sizes. As shown in prior works, in \emph{offline learning} (left), higher SGD noise (lower batch size) offers an implicit bias advantage and plateaus at a lower loss.  In contrast, we show that in \emph{online learning} (right), higher SGD noise does not provide any implicit bias benefit to performance, and lower noise reaches a smaller loss. The y-axis measures early stopping (true) loss.  See Section~\ref{sec:2} for more details.}
    \label{fig:intro}
\end{figure*}

Perhaps counter-intuitively,  SGD noise is often deemed \emph{advantageous} for implicit bias.
In particular, several works showed that higher learning rates and smaller batch sizes yield \textit{flatter} minima \citep{small-batch2017, LeCun1998, revisiting-bs, hour-imagenet}, which tend to generalize well \citep{Hochreiter1997,SmithED20} (see also Figure~\ref{fig:intro}, left).
However, these works are limited to the setting of multi-epoch or \emph{offline training}.

In this work, we examine the implicit bias of SGD in the \textbf{online learning} setting, in which data is processed through a \emph{single epoch}.
Online learning is common in several self-supervised settings, including large language models (LLMs)~\citep{komatsuzaki2019one,brown2020language,hoffmann2022an,hoffmann2022training,chowdhery2022palm}.
While in online learning, the train and test distributions are identical (and hence there are no  generalization considerations), it is still a \emph{non-convex} optimization. 
So, the inductive bias introduced by algorithm and hyperparameter choices could still potentially play a major role in learning trajectory and model quality.
However, we find that the impact of batch sizes in practical settings of online learning is qualitatively similar to their impact on \emph{convex} optimization. Specifically, we undertake an extensive empirical investigation and find that, \emph{in online learning, SGD noise is indeed only ``noise''} and offers no implicit benefits beyond optimization efficiency. 
This can be seen in Figure \ref{fig:intro} (right), where we observe that (neglecting computational cost) performance in online learning improves with increasing batch size.

 \begin{figure*}[!htb]
     \centering
     \includegraphics[width=4.5in]{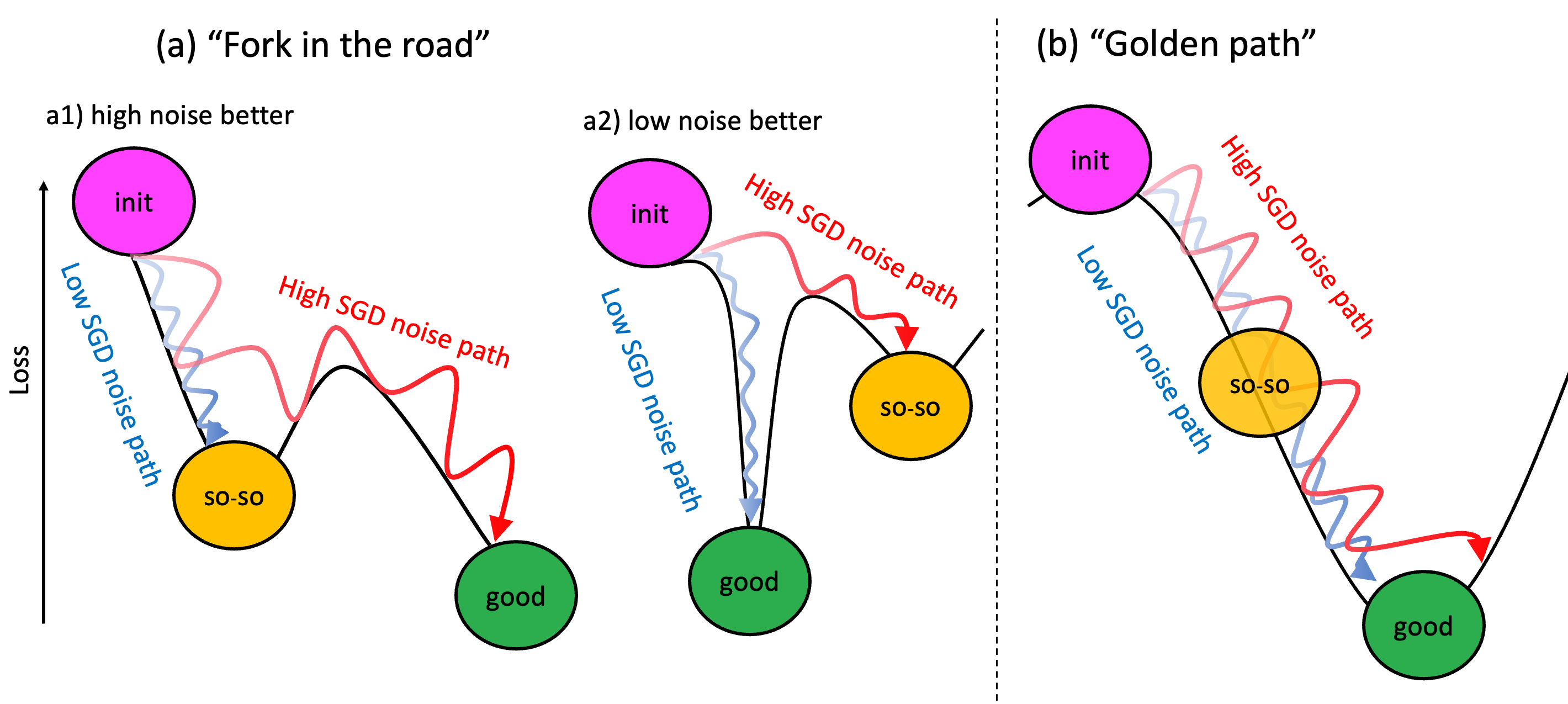}
     \caption{A-priori online learning can exhibit two potential scenarios: \textbf{(a)} \emph{``Fork in the Road,''} wherein the selection of batch size leads the optimization algorithm to explore distinct regions of the search space, potentially resulting in different loss outcomes.  (a1: better loss for the high-noise path, which is the common case for offline learning,  and a2: better loss for the low-noise path). \textbf{(b)} \emph{``Golden path,''} wherein the optimization trajectory remains similar for both gradient descent and SGD. In the latter scenario, the noise in SGD primarily influences the algorithm's traversal speed (and stability) along the path. Our research provides evidence supporting the ``golden path" scenario for online learning.\looseness=-1}
     \label{fig:cartoon}
 \end{figure*}

\textbf{The ``Golden Path'' Hypothesis}. When taking SGD to the limit of large batch sizes, we get the \emph{noiseless}  \emph{Gradient Descent} (GD) algorithm. Our findings hint that the SGD path is just a noisy version of the underlying noiseless GD path as illustrated in Figure~\ref{fig:cartoon}. We propose this conjecture as the \textit{golden path} hypothesis, which stands in contrast to the alternative ``Fork in the Road" possibility, wherein SGD and GD discover qualitatively different functions.
As shown in prior works (as well as in our own experiments), the ``fork in the road'' scenario (and specifically scenario a1, where high noise leads to a better minima) is the typical case for \emph{offline learning}.

To be precise, our work gives evidence for the following hypothesis:

\begin{center}
    
\fbox{
\begin{minipage}{0.9\linewidth}
\textbf{Golden Path Hypothesis}: For natural settings in online learning, a path trained via SGD does not deviate far from one trained via GD, in the following sense:  
\begin{itemize}
[itemsep=1.5pt,topsep=2pt,listparindent=2pt,leftmargin=20pt]
    \item \textbf{Loss Trajectories}: If SGD noise is dropped\footnotemark\;from high to low at time $t$, then shortly after $t$, the loss curve will ``snap’’ to track the curve of a model that was trained with low noise from initialization, and hence follow the ``golden path’’.
    \item \textbf{Function Space}: After reducing the SGD noise, the resulting function from this path is also similar in functional distance to that of a path with low noise from initialization.
\end{itemize}
\end{minipage}
}
\end{center}
\footnotetext{Note that high SGD noise runs have more variance, and to remove this factor we equalize the noise of two trajectories before comparison. See Figure~\ref{fig:cifar-5m:pointwise:highlr} for an example.}
\textbf{Is the Hypothesis False?} Indeed, the ``Golden Path'' hypothesis does not hold in a variety of settings. One simple toy example is provided in Figure 2 of \citet{Let18}, demonstrating how higher noise could help in escaping ``sharp'' minima in the non-convex regime (for further discussion, refer to the related works). However, for real world datasets trained using deep learning, we observe that the SGD path becomes ``close'' to that of the GD path once the SGD noise has been reduced. As we have stated above, closeness is measured both in terms of the loss trajectories and functional distance. One could also consider a stronger notion of the golden path hypothesis in weight space; however, in its naive formulation, the golden path hypothesis \textbf{does not hold in weight space}; this is due to the presence of permutation symmetries, dead neurons, and structural differences (see section~\ref{sec:related}). While we believe that a more careful formulation of the ``golden path’’ hypothesis can hold in weight space, we focus on loss and function space in this paper and defer weight space exploration for future work.

Overall, our work gives evidence to the hypothesis that there is a ``noiseless'' or ``golden'' path that Gradient Descent takes,  and that batch size plays no role in the choice of the path but only in the computational cost to travel on it, the training stability as well as the level of ``variance'' along the path. 
Hence choosing batch sizes should be determined by balancing their negative impact on noise with their positive impact on computation.
This is in stark contrast to the role of SGD in \emph{offline learning}, wherein SGD noise can influence not just the \emph{speed} of optimization but also its \emph{journey} and even its final \emph{destination} (i.e., function at convergence).

\textbf{Contributions and organization.} We delineate our contributions as follows:

\begin{enumerate}[itemsep=1.5pt,topsep=2pt,listparindent=2pt,leftmargin=20pt]
\item Our first contribution is demonstrating that, unlike in offline learning, SGD noise does not provide any implicit bias advantage in a variety of practical online learning settings. This is presented in Section~\ref{sec:2}, which contains a systematic investigation of the effects of SGD noise in the online versus offline settings. Our analysis encompasses both vision (ResNet-18 on CIFAR-5m, ConvNext-T on ImageNet) and language tasks (GPT-2-small on C4), 

\item A second contribution is to propose and examine the ``golden path'' hypothesis in the context of online learning. In Section~\ref{sec:snaploss}, we provide evidence that SGD loss trajectories follow that of gradient descent by showing that the loss curves of high-noise SGD ``snap'' to those of low-noise SGD when the noise levels are equalized.
\item In Section~\ref{sec:pointwise}, we further substantiate the ``golden path'' hypothesis in \emph{function space}.  We present evidence that models trained with varying levels of SGD noise learn similar functions, indicating that the differences in noise do not significantly impact the learned representations.
\end{enumerate}

Overall, our work sheds new light on the role of batch size in online deep learning, showcasing that its benefits are merely computational.
We also provide a pathway for a more unified understanding of training trajectories, by giving evidence that SGD takes noisy steps that approximate the ``golden path'' taken by gradient descent.

\subsection{Related Work}\label{sec:related}

\textbf{Implicit Bias:} A considerable volume of literature has been devoted to examining the impact of batch size on the training of neural networks from both theoretical \citep{andriushchenko2022sgd, catapult, control-bs-lr, pmlr-v162-nacson22a,PaquettePAP22,paquette2022homogenization} and practical  \citep{sgdnotneeded, nado-adam-suffices, revisiting-bs, xing2018walk, fisher-explosion2020, karpathy2019recipe} perspectives. Among practitioners, the consensus revolves around maximizing computational resources: large batch sizes are employed to fully exploit the hardware. However, regarding optimal hyperparameters, it is widely held that smaller batch sizes result in superior minima \citep{small-batch2017, LeCun1998, revisiting-bs, hour-imagenet}. Although some empirical studies \citep{sgdnotneeded, lee2022achieving, novack2022disentangling, hoffer2018train} contest this notion by utilizing various techniques, it remains the prevalent intuition within the community.
From a theoretical standpoint, several works showed the benefit of SGD noise as yielding a more favorable implicit bias~\citep{damian2021, haochen21a, ali2020implicit, blanc20a, Break-Even}. These works show that in certain overparameterized settings, higher SGD noise leads to a better generalization.

Within non-convex optimization literature, it is known that SGD can exhibit important optimization effects such as escaping saddle points and sharp minima \citep{Kleinberg18}. A simple example of such escape behavior is provided in Figure 2 of \citet{Let18}. 
In convex optimization, however, diminishing the stochastic gradient descent (SGD) noise typically leads to enhanced performance. For instance, \cite{PaquettePAP22, paquette2022homogenization} establish that, under certain assumptions for high-dimensional random features models, a reduced batch size results in a worse test error for a given number of training steps. These works operate within a regime where number of data points scales proportionally with the model size, thereby aligning more closely with the ``online learning'' paradigm. 

\textbf{Offline vs. Online:} One distinction absent from the aforementioned discussion is the difference between the online and offline regimes. For instance, \cite{SmithDBD21} clearly investigate the effect of SGD noise for a single epoch training, and show that it has an implicit bias towards reducing gradient norms. In contrast, we observe empirically that in the online setting, implicit bias doesn't affect the network in function space. The Deep Bootstrap framework of \cite{nakkirandeep, ghosh2021stages} contrasts the online and offline worlds, revealing that a significant portion of offline training gains can be attributed to its online component. Recent works also demonstrated the detrimental effects of repeating even a small fraction of data~\citep{hernandez2022scaling, xue2023repeat}, for LLMs. 

\textbf{Network Evolution.} Similar to us, multiple works discuss the similarity of SGD dynamics across hyperparameter choices. This question has been studied from the lenses of  example  order \citep{increasing, simplicity-bias, hacohen2020let, Behnam-difficulty}, representation similarity \citep{Kornblith0LH19, yamini-stitching}, model functionality  \citep{olsson2022incontext}, loss behaviour \citep{DDD}, weight space connectivity \citep{FortDPK0G20, pmlr-v119-frankle20a} and the structure of the Hessian \citep{CohenKLKT21}. Our work focuses on the online regime, and as opposed to previous studies, gives evidence to the \emph{Golden Path} conjecture in this regime wherein SGD noise strides (in function space) along the gradient descent trajectory but with noise. As shown by both our work and others, the golden path conjecture does \emph{not} hold for offline learning.
\section{The Implicit Bias of SGD in Online Learning} \label{sec:2}

\begin{figure*}[!ht]
    \centering
    \begin{subfigure}[t]{0.31\textwidth}
        \includegraphics[width=\textwidth]{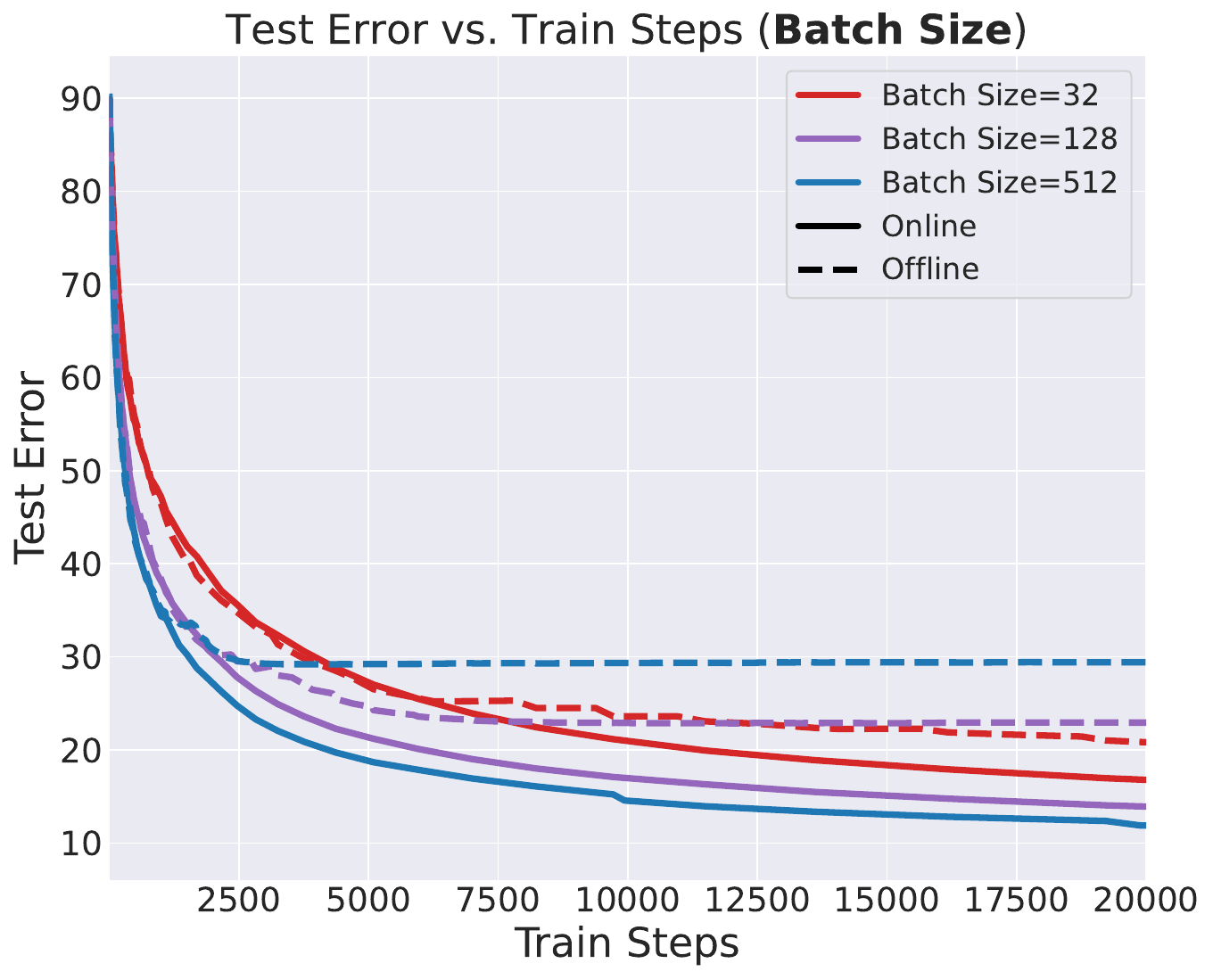}
        \caption{CIFAR-5m}
        \label{fig:lr-c5m}
    \end{subfigure}
    \hfill
    \begin{subfigure}[t]{0.32\textwidth}
        \includegraphics[width=\textwidth]{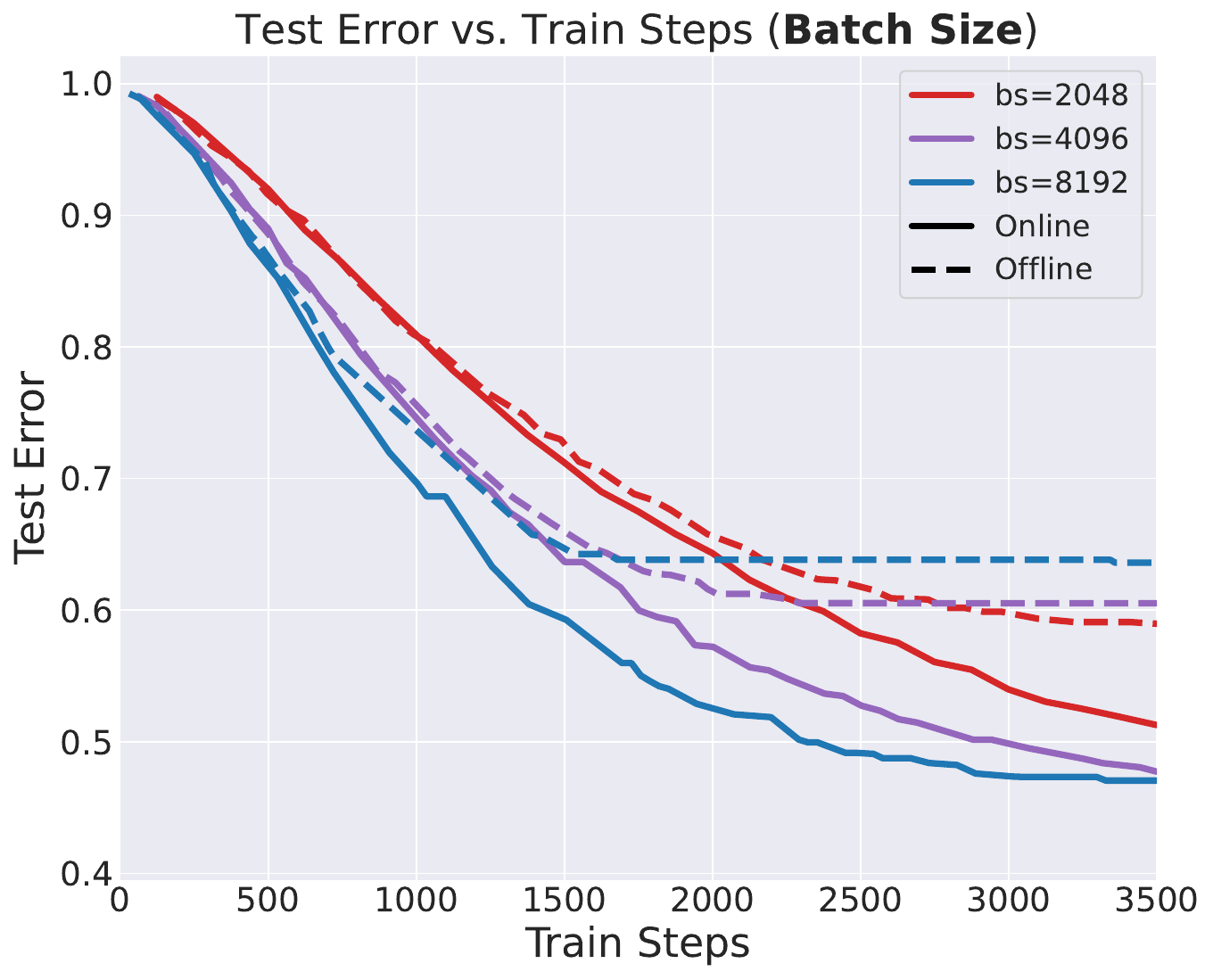}
        \caption{ImageNet}
        \label{fig:lr-imagenet}
    \end{subfigure}
    \hfill
    \begin{subfigure}[t]{0.31\textwidth}
        \includegraphics[width=\textwidth]{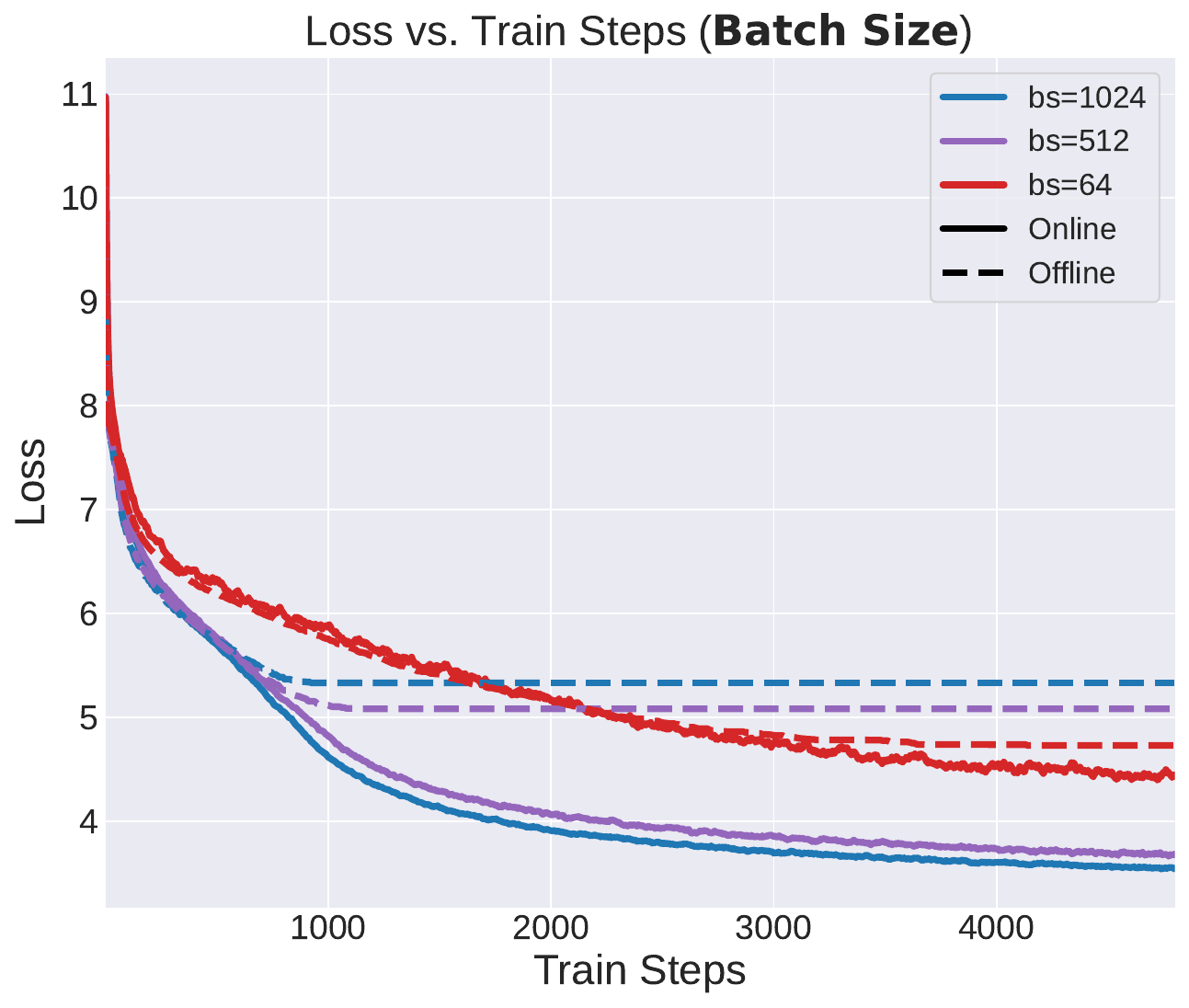}
        \caption{C4}
        \label{fig:lr-llm}
    \end{subfigure}
        \caption{Test performance for ResNet-18 trained on CIFAR-5m (\textbf{left}), ConvNext-T on ImageNet (\textbf{middle}), and GPT-2-small on C4 (\textbf{right}) across varying batch sizes. Red corresponds to high SGD noise (small batch size), blue to low SGD noise (high batch size), and purple to an intermediate setting. Solid (resp. dotted) lines correspond to runs in the online (resp. offline) setting. For online learning, lower SGD noise runs consistently outperform higher noise runs per given step. Offline learning performance initially matches online performance, eventually runs with higher noise  outperform low-noise runs. All experiments are averaged over $\geq 4$ runs. See Figure~\ref{fig:c4_multiple_seeds},~\ref{fig:cifar_multiple_seeds} for error bars and more hyperparameter values.}
    \label{fig:lrbatch}
\end{figure*}

In this section, we present our experimental results on the impact of SGD noise (i.e. batch size) on implicit bias. We show that the effect of this noise can differ significantly between the \emph{offline} and \emph{online} regimes. 
Since our goal is to study the impact of SGD noise on implicit bias rather than on computational efficiency, in our experiments, we measure loss as a function of the number of gradient steps, and not as function of datapoints seen.

We conduct an experimental evaluation of our claims employing convolutional models in computer vision and Transformer models in natural language processing (NLP). Specifically, we run ResNet-18 on CIFAR-5m \citep{nakkirandeep}, a synthetically generated version of CIFAR-10 with 5 million examples, ConvNext-T on ImageNet, and GPT-2-small on C4. To imitate the online regime with ImageNet, we only train for 10 epochs with data augmentation. For full experimental details, refer to Appendix~\ref{app:experiments}. As we show in Figure~\ref{fig:lrbatch}, we find that

\begin{enumerate}[itemsep=1.5pt,topsep=2pt,listparindent=2pt,leftmargin=20pt]
    \item In the \textbf{offline setting}, consistent with prior work~\citep{small-batch2017}, SGD noise can (and often does) lead to better implicit bias for the final models. Specifically, even if runs with smaller noise initially\footnote{The curves for offline learning initially track the online learning curves (as predicted by Deep Bootstrap~\citep{nakkirandeep}) but then plateau at a higher loss for lower SGD noise.} decrease the loss faster, eventually they get ``stuck'' at a worse local minima than the runs with higher SGD noise (smaller batch size). This is consistent with Scenario a1 of Figure~\ref{fig:cartoon} (``fork in the road'' with high noise being better), where a higher noise enables escaping from bad local minima. 
    \item In contrast, in the \textbf{online setting}, the implicit bias advantage of SGD noise \emph{completely disappears}, and the main benefit from small batch sizes reduces to being just computational. Specifically, after we control for computation (by measuring number of gradient steps), the low-noise runs consistently outperform the higher noise runs. This is consistent with either Scenario a2 of Figure~\ref{fig:cartoon} (``fork in the road'' where a lower noise run can explore better minima) or with Scenario b (the ``golden path'': higher noise follow a similar trajectory but with some degradation due to noise). 

    First, we theoretically show, that for convex online setting, in particular quadratic loss, SGD path always remains close to the GD path in expectation, and SGD noise simply acts as variance on top of this path, thus supporting the golden path hypothesis (these results are similar in spirit to the results of \citet{PaquettePAP22}).
    
    \begin{theorem} \label{thm:sgd:gd:convex}
    Consider the quadratic loss function given by $\mathcal{L}(w) = w^\top Hw$, where $H$ is a positive semi-definite matrix. With stochastic gradients (denoted by $g(w)$) modeled as additive gaussian noise, i.e, $g(w) = \nabla \mathcal{L}(w) + \xi$, where $\xi \sim \mathcal{N}(0, \sigma^2 I)$, and for a fixed learning rate schedule, the following holds:
    \begin{itemize}
        \item Consider two different SGD runs, $R_1$ and $R_2$, starting from the same initialization $w_0$ and having noise variances given by $\sigma_1$ and $\sigma_2$ ($\sigma_1 \geq \sigma_2$). Then $\E[\mathcal{L}(w_{R_1}(t))] \geq \E[\mathcal{L}(w_{R_2}(t))]$.
        \item Let the noiseless GD run from the same initialization $w_0$ be denoted by $w_{GD}(t)$. Then $\E[w_{R_1}(t)] = \E[w_{R_2}(t)] = w_{GD}(t)$.
    \end{itemize}
    \end{theorem}
    
    Moreover, in Sections~\ref{sec:snaploss} and~\ref{sec:pointwise}, our additional experimental results give evidence for the ``golden path'' case in practical scenarios.
\end{enumerate}

%Concretely, we conduct online and offline training experiments and report the loss (or error) as a function of the number of steps for the batch size experiments and LR-normalized steps for the learning rate ones. We observe qualitatively similar effects of a higher learning rate and a smaller batch size and  demonstrate that for online learning, having more SGD noise only results in higher loss. 
%To conclude, in the online learning settings we consider, SGD noise offers no implicit-bias benefit.
%In contrast, the curves for offline learning initially track the online learning curves (as predicted by Deep Bootstrap~\citep{nakkirandeep}) but then plateau at a higher loss for lower SGD noise. %Our experiments suggest that the low-noise regimes we explore are close approximations of \emph{Gradient Flow}. 

\textbf{Isn't this trivial?} One natural intuition is that regularization is not needed in the online learning setup because there is no difference between optimization and generalization (or between train and test). However, we have empirically observed that well-known explicit regularization techniques such as weight decay can yield better performance in the online regime for standard tasks (Figure~\ref{fig:prob1_6_1}, Appendix~\ref{app:add-2}), suggesting that this intuition might be flawed and more nuance is required.

\section{Snapping Back to the Golden Path} \label{sec:snaploss}
\begin{figure*}[!ht]
  \centering
  \begin{subfigure}{0.42\textwidth} \hspace{-0.5cm}\includegraphics[width=\linewidth]{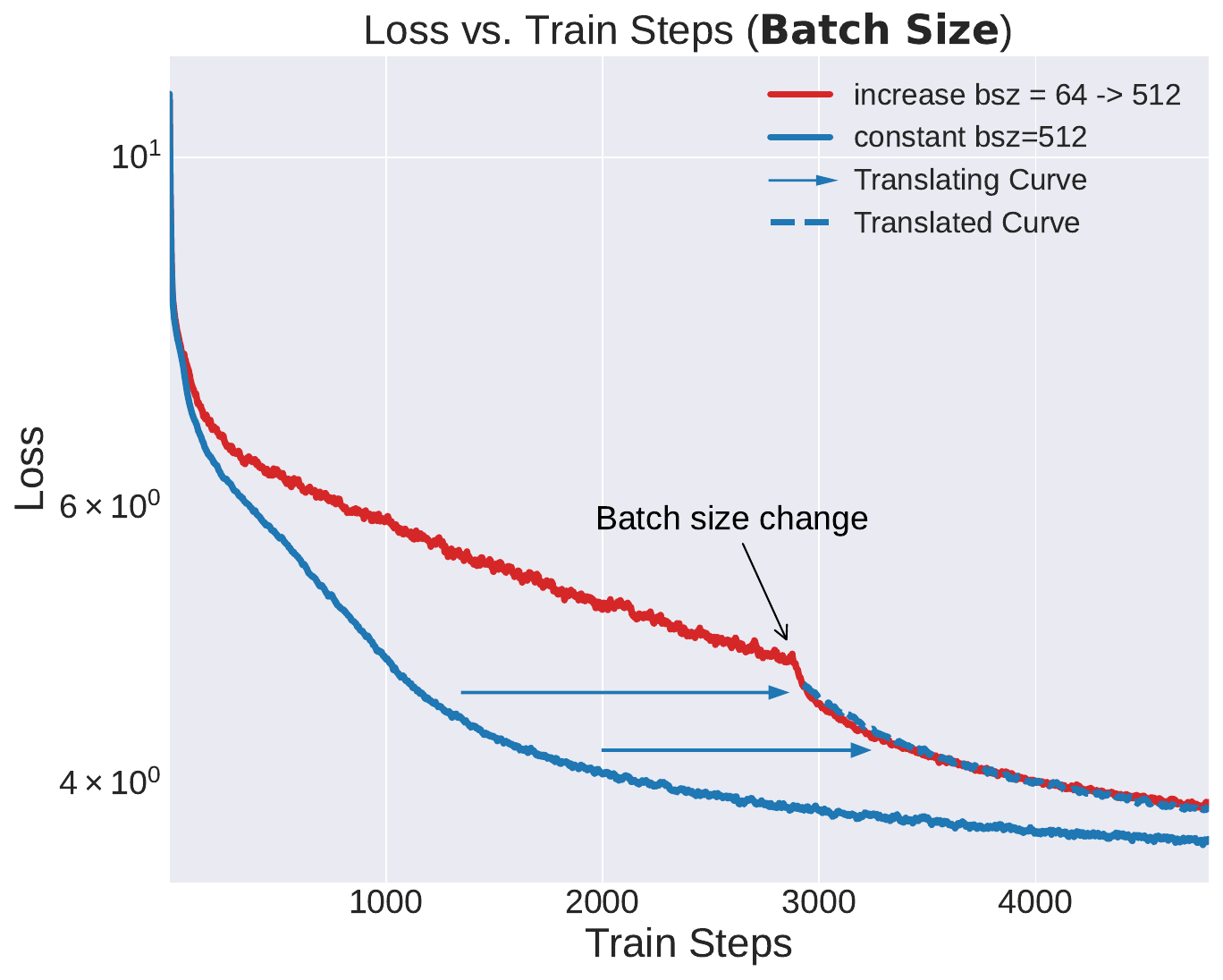}
    \caption{Increasing Batch Size: C4}
    \label{fig:c4-inc-bsz}
  \end{subfigure}%
  \quad\quad\quad
  \begin{subfigure}[b]{0.42\textwidth}\includegraphics[width=\textwidth]{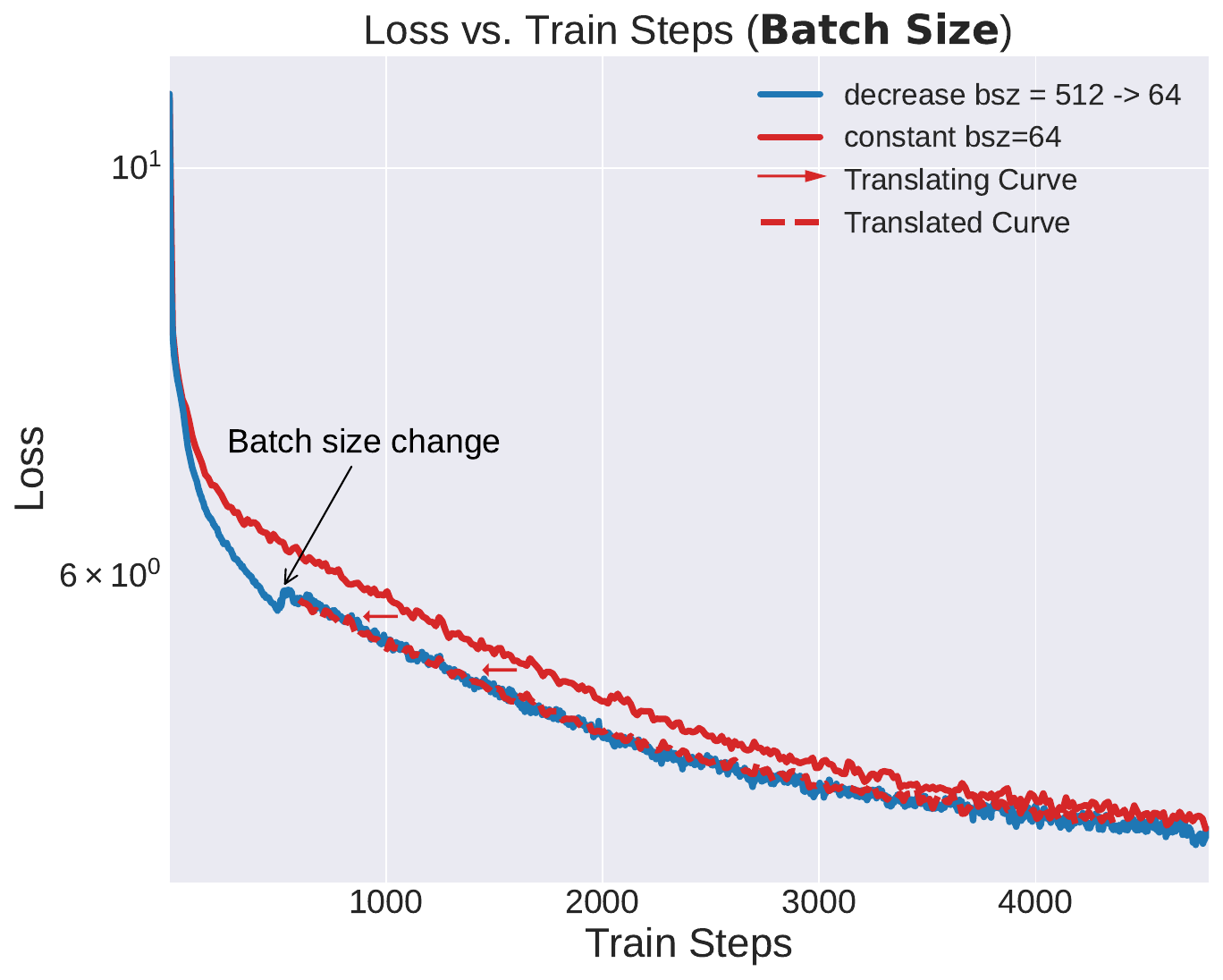}
    \caption{Decreasing Batch Size: C4}
    \label{fig:c4-dec-bsz}
    
  \end{subfigure}

  \begin{subfigure}{0.42\textwidth}
    \includegraphics[width=\linewidth]{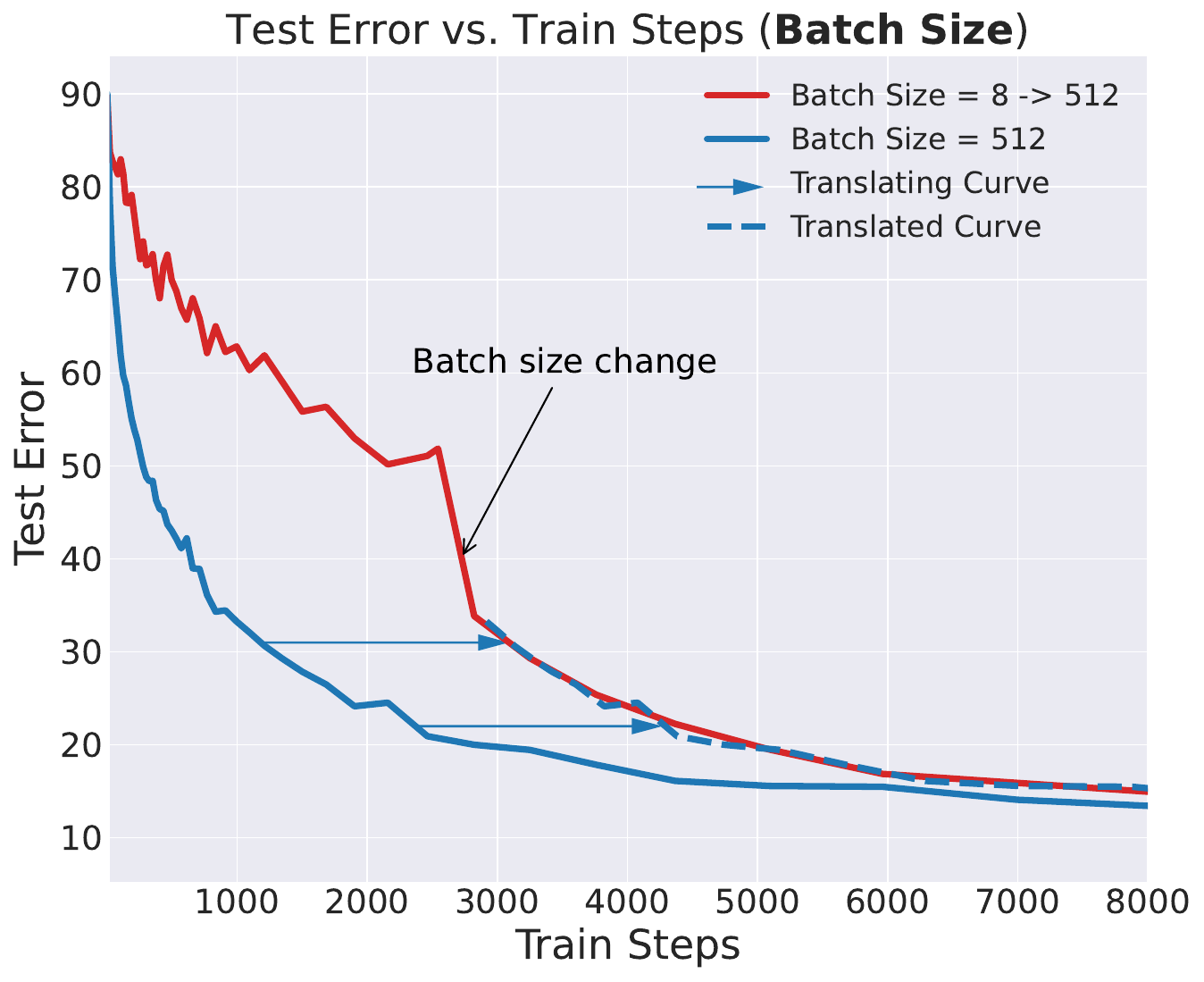}
    \caption{Increasing Batch Size: CIFAR 5m}
    \label{fig:cifar-inc-bsz}
  \end{subfigure}%
  \quad\quad\quad\quad
  \begin{subfigure}{0.42\textwidth}
    \includegraphics[width=\linewidth]{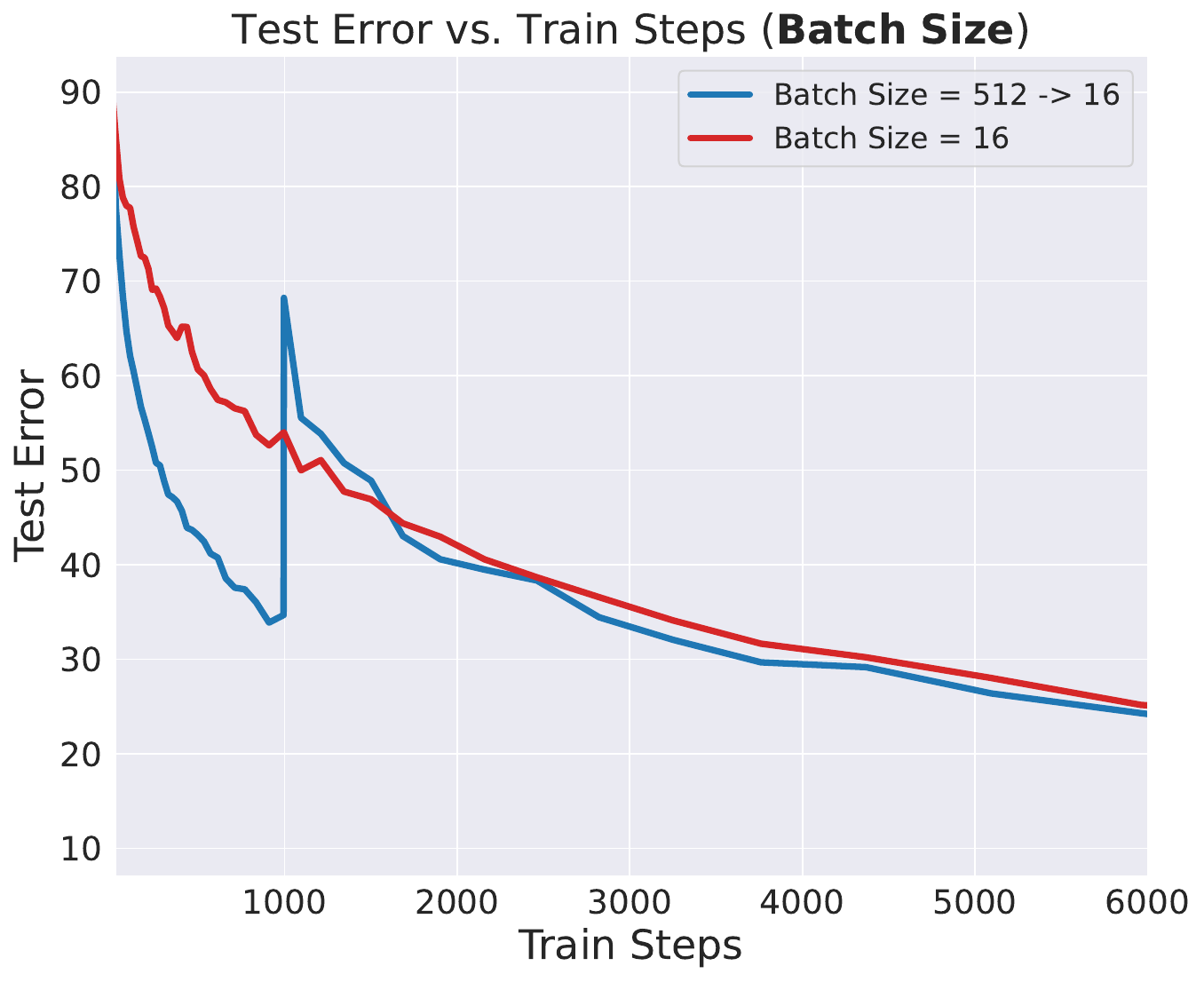}
    \caption{Decreasing Batch Size: CIFAR 5m}
    \label{fig:cifar-dec-lr}
  \end{subfigure}

  \caption{Changing SGD noise (\textbf{left}: increasing batch size, \textbf{right}: decreasing batch size) during training for ResNet-18 on CIFAR-5m (bottom) and GPT2-small on C4 dataset (top). The red curves correspond to models trained with high SGD noise from initialization, and the blue curves trained with low SGD noise from initialization. In left plot the batch size is increased after $T_0$ steps while in right plot the batch size is decreased after $T_0$ steps. Across both experiments, changing batch size causes the original curve to follow a translated version (dashed) of new batch size curve (except for increasing batch size experiment on CIFAR-5m where no translation is required).}
  \label{fig:sec2-main}
\end{figure*}

The results of Section~\ref{sec:2} show that SGD noise does not provide beneficial implicit bias in online learning.
As we discussed in Figure~\ref{fig:cartoon}, there are two potential explanations for why in online learning (unlike the offline case), decreasing SGD noise steers optimization towards a smaller-loss trajectory.
One explanation is Scenario (a2) of the figure.
Namely, it may be the case that choosing low SGD noise leads the optimization algorithm to a different (and better) trajectory, that is completely inaccessible to the high SGD-noise runs. 
The second is the ``golden path'' hypothesis: higher-noise runs travel on approximately the same path as lower-noise ones, suffering some loss-degradation resulting from the imperfect approximation.
To rule out the first explanation, we conduct the following experiment in the online setting (see Figure~\ref{fig:sec2-main}, left):
\begin{enumerate}[itemsep=1.5pt,topsep=2pt,listparindent=2pt,leftmargin=20pt]
    \item Run two experiments---one with high batch size, and one with small batch size---for $T_0$ steps.
    \item After $T_0$ steps, decrease the SGD noise by increasing the batch size of the second experiment to match the hyperparameters of the first one, and continue both runs.  
\end{enumerate}
Under the golden path hypothesis, we expect that  shortly after increasing the batch size (i.e., at $T_0 + \tau$ for $\tau \ll T_0$), the loss curve would ``snap'' to the golden path, and continue following the same trajectory of the model that was trained with low SGD noise. On the other hand, the (a2) scenario of Figure~\ref{fig:cartoon} implies that decreasing the noise would not result in any significant change to the loss curve. 

We perform a series of experiments as described above with ResNet-18 on CIFAR-5m and GPT2-small on the C4 dataset in Figure~\ref{fig:sec2-main} ((a) and (c)). We consistently observe that, after dropping the SGD noise at some time $T_0$, the loss sharply improves to some value $\ell_0$. From this point onward, the loss curve of the model is nearly identical to a right-translation of the loss curve of the model that was trained with low SGD noise from initialization. 
This phenomenon does not hold generally in offline learning, as it is well known that at convergence, different minima are reached by gradient descent and SGD.\looseness=-1

In contrast to exploring the transition from a noisy trajectory towards the ``golden path'', we can also investigate the effects of introducing  SGD noise, thereby deviating away from the optimal trajectory. 
To this end, we conduct an analogous experiment to the one presented before, but with a crucial difference: instead of reducing the noise in SGD at time $T_0$, we \textit{increase} it by decreasing the batch size. Looking at the results for ResNet-18 on CIFAR-5m and GPT-2-small on the C4 dataset, we observe an immediate and significant increase in the loss upon introducing additional noise, as illustrated in Figure~\ref{fig:sec2-main} ((b) and (d)). We see a similar phenomenon as in the experiments shown in Figure~\ref{fig:sec2-main} (left), where the lower noise loss curve, after noising, follows a translated version of the higher noise curve (translation is not needed for the CIFAR-5m experiment as shown in Figure~\ref{fig:sec2-main} bottom right).

In Appendix~\ref{app:add-3}, we provide additional plots showing that the phenomenon is consistently observed even when varying the timestep when the batch size is increased.
\begin{figure*}[!ht]
    \centering
    \includegraphics[width=5in]{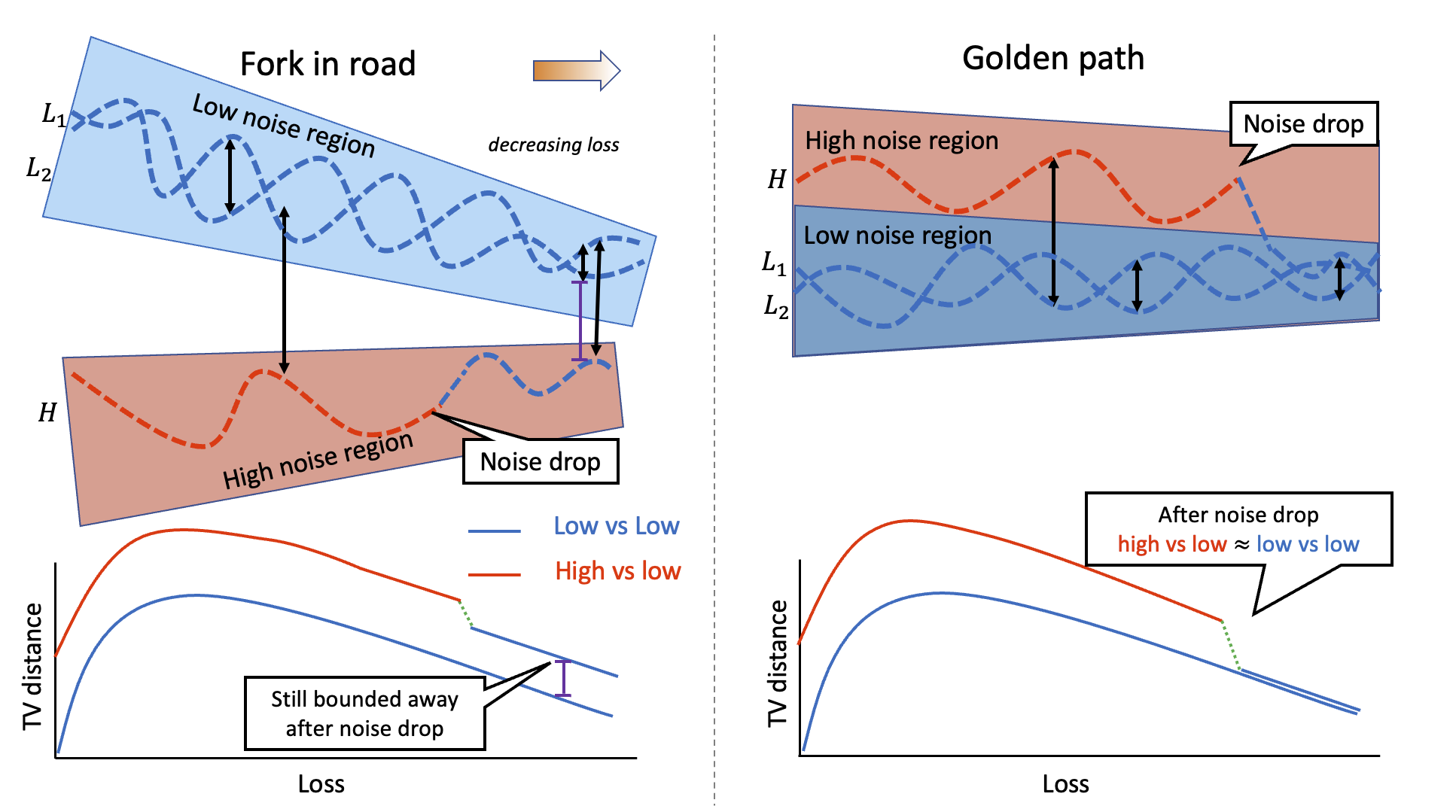}
    \caption{
    This figure illustrates the potential total variation distance in function space for two scenarios: ``fork in the road" (left) and ``golden path" (right). 
In the ``fork in the road" scenario, training runs with low and high batch sizes explore different regions of the function space, leading to a consistently high total variation distance, even when the batch size is increased. In the "golden path" scenario, the low batch size run follows a noisy version of the high batch size trajectory. Increasing the batch size causes the trajectory to align with the high batch size path, resulting in a total variation distance similar to two independent high batch size trajectories.
}
    \label{fig:cartoon-distance}
\end{figure*}

\section{Pointwise Prediction Differences Between Trajectories} \label{sec:pointwise}

Section~\ref{sec:snaploss} offered empirical substantiation for the golden path hypothesis, but was restricted to claims about the \emph{loss curve}. Now we investigate a stronger version of this hypothesis: the golden path in \textbf{Function Space.} Specifically, our goal is to verify whether trajectories using different SGD noise are functionally similar. Due to differing noise levels between the trajectories, we cannot directly compare their functional similarity; instead, we show that lowering the SGD noise causes models to ``snap'' to the golden path in a functional sense, i.e. when measuring the \emph{pointwise} distance.

\begin{figure*}[!ht]
    \centering  
    \begin{subfigure}[b]{0.45\textwidth}
    {\includegraphics[width=\textwidth]{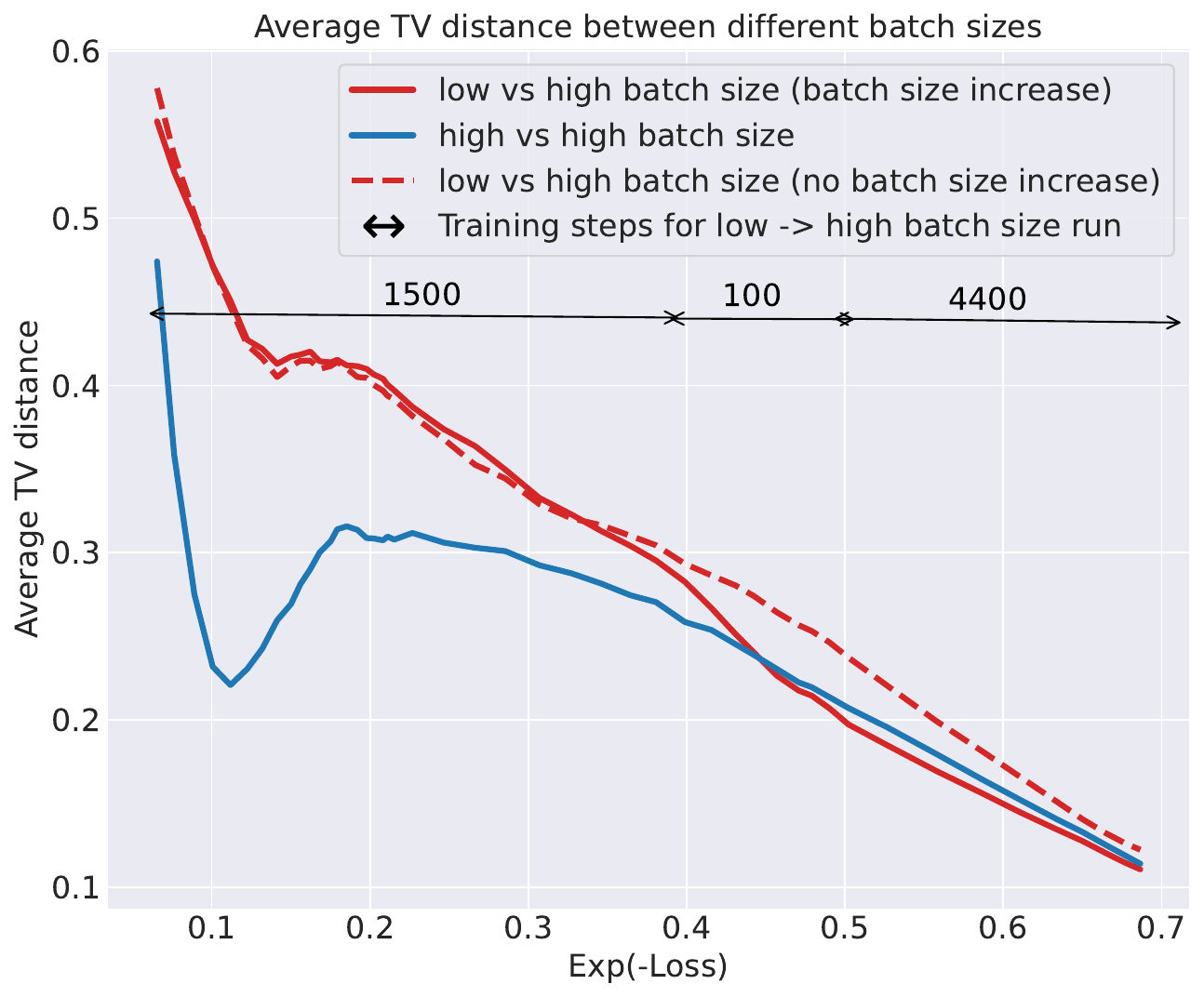}\caption{CIFAR-5m}}
    \end{subfigure}
    \quad\quad\quad\quad
    \begin{subfigure}[b]{0.45\textwidth}
    {\includegraphics[width=\textwidth]{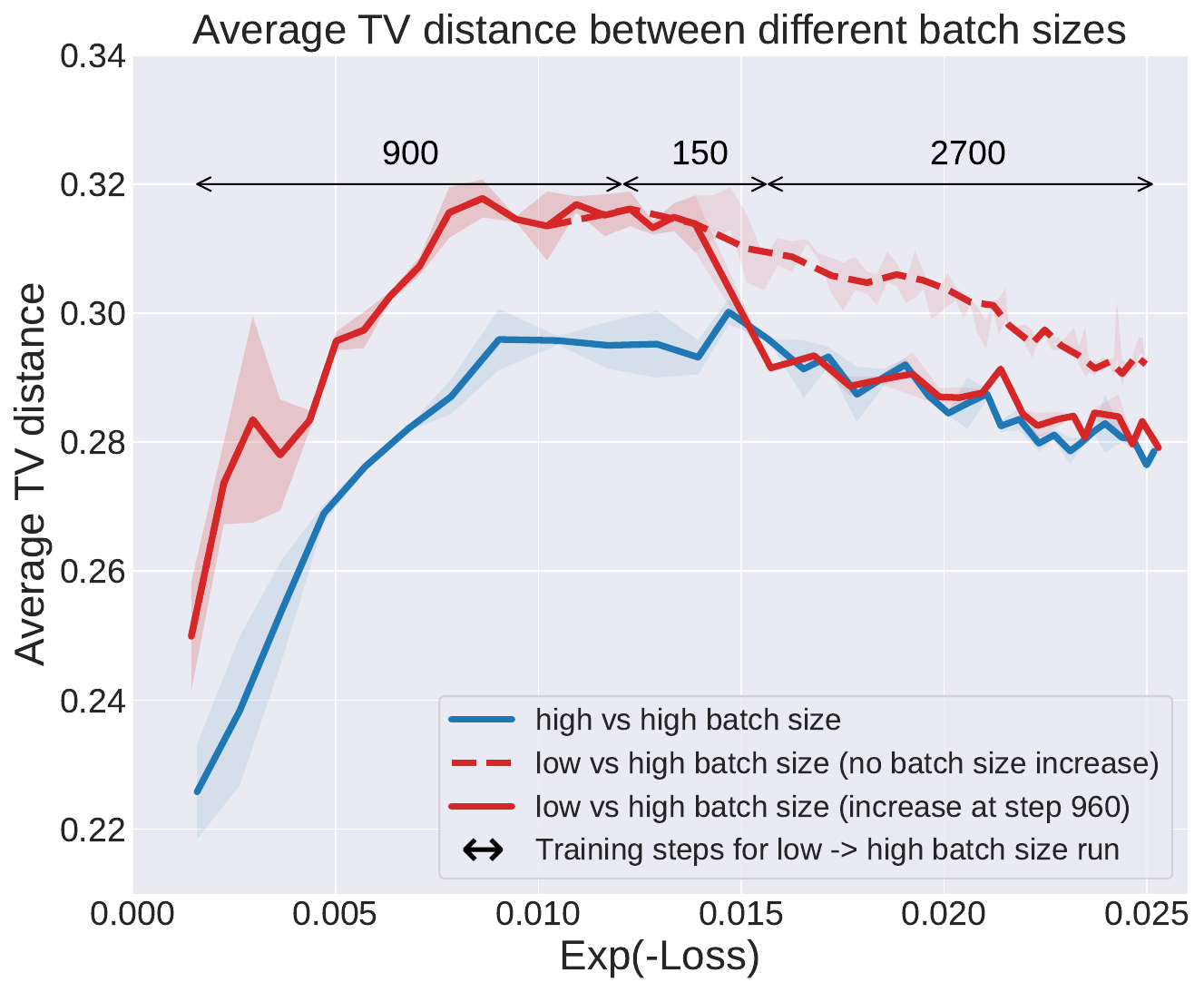}\caption{C4} \label{fig:bsz_functional}}
    \end{subfigure}
  \caption{Average TV distance behavior on CIFAR-5m (left) and C4 (right) with and without reducing SGD noise. For CIFAR-5m, we increase the batch size from 32 to 512. For C4, we increase the batch size from 64 to 512. The dashed line represents the run with constant high SGD noise as a baseline. The figure demonstrates that the distance between the model with reduced noise and a low-noise model at the same loss is nearly identical to the distance between two random low-noise models trained with the same hyperparameters. This is in contrast with the distance between a high noise and a low-noise model at the same loss. This observation supports the ``golden path hypothesis'' and contradicts the ``fork in the road'' hypothesis. }
  \label{fig:cifar-5m:pointwise:highlr}
  \end{figure*}

To be precise, we investigate the functional distance between models by measuring the average total variation (TV) distance\footnote{The TV distance is defined as half of the $\ell_1$ distance for probability measures: $TV(p, q) = \frac{1}{2} \sum_i |p_i - q_i|$.} of their softmax probabilities on the test dataset. Figure~\ref{fig:cartoon-distance} gives a schematic illustration of our experimental setting. In this model, we have a low noise SGD run to serve as the ``ground truth'' for the golden path. The blue curve in Figure~\ref{fig:cartoon-distance} serves as a baseline for the functional distance (TV) to this ``golden path'' as it represents another low noise run trained from a different initialization. This is a strong baseline since different initialization seeds are often treated as a nuisance parameter in deep learning. We compare it to the TV distance of the high noise path depicted by the red curve. 

As illustrated in Figure~\ref{fig:cartoon-distance}, under the ``fork in the road'' hypothesis, we expect that high noise and low noise trajectories will explore different regions of the search space. Thus, even if dropping the noise improves loss, the baseline (distance between independent runs of low SGD noise) will still be significantly lower than the distance between the `dropped-noise model' and the low-noise model.

In this section, by an empirical study on CIFAR-5m and C4 dataset, we show that this is not the case, and TV distance behaves as would be expected from the ``golden path hypothesis''. Specifically, as shown in Figure \ref{fig:cifar-5m:pointwise:highlr}, after dropping the noise, the TV distance between this model and a low noise model at the same loss is virtually identical to the TV distance between \emph{two random low-noise models trained with identical hyper parameters}. Figure \ref{fig:cifar-5m:pointwise:highlr} also shows that this is not true for the high noise model, demonstrating that the two paths indeed differ due to the difference in the level of noise. Moreover, in Figure \ref{fig:cifar-5m:pointwise:highlr}, it could also be seen that the high noise model reaches the ``golden path'' very quickly upon dropping the SGD noise. In Appendix~\ref{app:add-4}, we provide additional plots showing that this phenomenon holds across multiple different timesteps when the batch size is increased.

\section{Discussion and Conclusion}

The results of our investigation reveal a striking discrepancy between the online and offline learning regimes, in terms of the implicit bias of SGD. In the offline regime, this bias exhibits a beneficial regularization effect, whereas in the online regime, it merely introduces noise to the optimization path. This critical distinction between the two regimes has largely been overlooked in both theoretical and empirical research, with few studies explicitly addressing the difference. We argue that recognizing and accounting for the online versus offline learning regimes is crucial for understanding various deep learning phenomena and for informing the design of optimization algorithms.

Our experiments cover three practical datasets spanning vision and language, but further research may be needed to fully understand the generalizability of our findings. Although our work represents only an initial exploration into the disparities between the online and offline learning regimes, we can draw several immediate conclusions.

\textbf{Implications for Practitioners.} For online learning, our findings emphasize the relative \textit{simplicity} of hyperparameter tuning, primarily focusing on computational efficiency and stability.
In situations where data or computational resources are limited, however, the regularization effects of SGD become more significant, and hyperparameter selection and optimization take on greater importance. For instance, in the low-data regime, it may be crucial to use a smaller batch size, even if it results in not fully utilizing the GPU. In online training, this consideration appears to be consistently irrelevant.\looseness=-1

\textbf{Implications for Theorists.} The primary takeaway for theoretical research is that the study of the regularizing effects of high SGD noise should be confined to the offline learning regime, as failing to make this distinction creates tension with practical applications. Theoretical findings that do not account for this difference can not fully capture why SGD is effective for deep learning.
Furthermore, our observation that SGD follows a noisy trajectory near the ``golden path'' of gradient descent in loss and function spaces, coupled with the Deep Bootstrap \citep{nakkirandeep} assertion that a substantial portion of offline training can be explained by the online regime, implies that gradient descent may be instrumental in understanding many aspects of deep learning.

In conclusion, given that many large-scale deep learning systems, such as Language Models (LLMs), predominantly operate within the online learning regime, our findings challenge the conventional understanding of deep learning, primarily based on offline learning. We contend that it is necessary to reevaluate our comprehension of various deep learning phenomena in the context of online settings. Moreover, we propose gradient descent as a promising theoretical tool for studying online learning, considering the minimal impact of SGD noise on the functional trajectory.

\section*{Acknowledgements}

We thank Preetum Nakkiran, Ben L. Edelman and Eran Malach for helpful comments on the draft.

NV, DM, RZ, GK and BB are supported by a Simons Investigator Fellowship, NSF grant DMS-2134157, DARPA grant W911NF2010021,and DOE grant DE-SC0022199. This work has been made possible in part by a gift from the Chan Zuckerberg Initiative Foundation to establish the Kempner Institute for the Study of Natural and Artificial Intelligence.
SK, DM, and RZ acknowledge funding from the Office of Naval Research under award N00014-22-1-2377 and the National Science Foundation Grant under award \#CCF-2212841.

\section*{Impact Statement}
This paper presents work whose goal is to advance the field of Machine Learning. There are many potential societal consequences of our work, none which we feel must be specifically highlighted here.

\newpage

\bibliography{ref}
\bibliographystyle{icml2024}

%%%%%%%%%%%%%%%%%%%%%%%%%%%%%%%%%%%%%%%%%%%%%%%%%%%%%%%%%%%%%%%%%%%%%%%%%%%%%%%
%%%%%%%%%%%%%%%%%%%%%%%%%%%%%%%%%%%%%%%%%%%%%%%%%%%%%%%%%%%%%%%%%%%%%%%%%%%%%%%
% APPENDIX
%%%%%%%%%%%%%%%%%%%%%%%%%%%%%%%%%%%%%%%%%%%%%%%%%%%%%%%%%%%%%%%%%%%%%%%%%%%%%%%
%%%%%%%%%%%%%%%%%%%%%%%%%%%%%%%%%%%%%%%%%%%%%%%%%%%%%%%%%%%%%%%%%%%%%%%%%%%%%%%
\newpage
\appendix
\onecolumn

\newpage

\section{Experimental Details}\label{app:experiments}

\subsection{CIFAR-5m}

In our CIFAR-5m experiments, we trained ResNet-18, on normalized (across channels) images and using the SGD optimizer with 0.9 momentum. 

\paragraph{Section 2:} For Figure~\ref{fig:lrbatch} (a) we trained without any data augmentations. For both offline and online learning, we used a learning rate of 0.025. For offline learning we trained on a random subset of 50k samples (class balanced). We also used exponential moving average with coefficient of .8.
%\paragraph{Section 2:} For Figure~\ref{fig:lrbatch} (a) we trained without any data augmentations. For Figure~\ref{fig:lrbatch} (a, top) we used a learning rate of 0.025 and for Figure~\ref{fig:lrbatch} (a, bottom) we used a batch size of 512. For offline learning we trained on a random subset of 50k samples (class balanced). Both plots use exponential moving average with coefficient of .8.

\paragraph{Section 3:} For Figure~\ref{fig:sec2-main} (c,d) we trained with standard CIFAR data augmentation of a random crop (\texttt{RandomCrop(32, padding=4)} in Pytorch) and horizontal flip (\texttt{RandomHorizontalFlip()} in PyTorch), and used a learning rate of 0.025.

\paragraph{Section 4:} For Figure \ref{fig:cifar-5m:pointwise:highlr} (a), we trained on CIFAR-5m without any augmentations so as to remove the pointwise difference due to different augmentations. We trained two networks with learning rate 0.05, one with batch size 32 and the other with 512, both for 12000 steps ($\sim 1$ epoch for batch size 512) and the one with batch size 32 was changed to 512, after 1500 training steps.

\subsection{ImageNet}
For the ImageNet experiments, we used ConvNext-T \cite{convnext} and unless specified otherwise, used a batch size of 2048 and learning rate of $1e$-$4$ with the AdamW optimizer with weight decay $0.005$. For all experiments, we use cosine decay scaling of the learning rate with respect to training steps (not epochs). We used the \texttt{RandomHorizontalFlip} and \texttt{RandomCrop} augmentations and also preprocessed the dataset to be resized to $256 \times 256$ using OpenCV before training for speed purposes. For the offline results, we downsample the dataset by a factor of 10, i.e., use 128k examples. 

\subsection{C4}
For all experiments we trained GPT-2-small (124m parameters) on the C4 dataset with sequence length 2048. The optimizer we use is Adam without weight decay and a constant learning rate of $6 \times 10^{-4}$.

\paragraph{Section 2:} For Figure~\ref{fig:lrbatch} (\textbf{c}) we used a learning rate of $6 \times 10^{-4}$. For offline learning we trained on a random subset of roughly 100 million tokens, and all hyperparameters are otherwise identical with the analogous run in the online setting.

\paragraph{Section 3:} In Figures \ref{fig:c4-inc-bsz} and \ref{fig:c4-dec-bsz}, we change the batch size from $64$ to $512$ and vice versa. When changing the batch size we keep the learning rate fixed to $6 \times 10^{-4}$. For Figure~\ref{fig:c4-inc-bsz}, the change occurs after 60\% of the training duration has elapsed (i.e. after 2880 steps) and for Figure~\ref{fig:c4-dec-bsz} the change occurs after 960 steps.

\paragraph{Section 4:} In Figure~\ref{fig:bsz_functional} we plot the average TV distance averaged across three pairs of different seeds for the (high vs high) and (low vs high) batch size runs. The average TV distance is computed over 500 batches of size 64 (32000 samples). The plotting begins after 50 training steps have elapsed until the end of the training duration.

%\section{Additional Related Work}\label{app:related}

\section{Discussion of Results in Section 2}
In Figure \ref{fig:lrbatch}, we showed the stark contrast in offline vs online learning regarding their interaction with SGD noise. For the offline setting, higher SGD noise generally leads to a better performance, while in online learning, SGD noise only hurts performance.

However, upon closer observation, we can see that in all the plots of Figure \ref{fig:lrbatch}, offline learning performance closely follows the online learning performance throughout the majority of the training period before reaching a plateau. This is in agreement with the results of \cite{nakkirandeep}. They empirically demonstrated that, across a variety of scenarios, a major part of offline training can be explained away by online learning. 

This result, combined with our observation about the SGD trajectory being a noisy version of the ``golden path'' exhibited by gradient descent in online learning, shows that gradient descent is a useful theoretical tool even for studying offline learning. In particular, given that in practice, we choose hyperparameters to maximize test performance, this means that we move as far along the online trajectory as possible. The ``denoised'' version of this online trajectory is given by gradient descent on the population loss. Thus, our work proposes studying ``gradient descent on population loss'' as an alternative (or a ``denoised'' version) of studying SGD in offline learning.
% \section{Additional Experiments for Section 2: Convergence to Gradient Flow}\label{app:gradient-flow}
% \rz{I don't think we have a reference to this section in the main paper anymore? It seems to be commented out.}

% \nv{Large batch size plots which show that that convergence to gradient flow improves with width (for online), for offline cite EOS paper.}
%\section{Additional Experiments for Section 3}\label{app:sec3}
\section{Discussion of Results in Section 3}\label{app:sec3}
As exhibited in Figure \ref{fig:sec2-main}, we observe the loss snapping phenomenon even when we increase SGD noise during training. Intuitively, the model that initially has a low amount of SGD noise (high batch size) has progressed ``far along'' the path, and thus the additional noise results in progress lost on the path and an increase in loss. One plausible reason for this could be the recent phenomenon of Edge-of-Stability \citep{CohenKLKT21}, where the authors showed that SGD dynamically leads the model to an ``edge-of-stability'' curvature, i.e, increasing SGD noise  leads to an increase in loss in the successive steps. However, our paper shows that, following this brief increase in loss, the model is able to recover and continue training, albeit with higher SGD noise; thus the performance follows that of a noisier trajectory. 

\section{Proof of Theorem \ref{thm:sgd:gd:convex}}

\begin{theorem*}
    Consider the quadratic loss function given by $\mathcal{L}(w) = w^\top Hw$, where $H$ is a positive semi-definite matrix. With stochastic gradients ($g(w)$) modeled as additive gaussian noise, i.e, $g(w) = \nabla \mathcal{L}(w) + \xi$, where $\xi \sim \mathcal{N}(0, \sigma^2 I)$, and for a fixed learning rate schedule, the following holds:
    \begin{itemize}
        \item Consider two different SGD runs, $R_1$ and $R_2$, starting from the same initialization $w_0$ and having noise variances given by $\sigma_1$ and $\sigma_2$ ($\sigma_1 \geq \sigma_2$). Then $\E[\mathcal{L}(w_{R_1}(t))] \geq \E[\mathcal{L}(w_{R_2}(t))]$.
        \item Let the noiseless GD run from the same initialization $w_0$ be denoted by $w_{GD}(t)$. Then $\E[w_{R_1}(t)] = \E[w_{R_2}(t)] = w_{GD}(t)$.
    \end{itemize}
    \end{theorem*}

\begin{proof}
    Consider the stochastic gradient descent update given by
    \[ w(t+1) = w(t) - \eta(t) g(w(t)) = (I - \eta(t)H)w(t) - \eta(t)\xi(t) \]
    Unrolling the update, we get
    \[ w(t+1) = \left[\Pi_{i=0}^t (I - \eta(i)H)\right] w(0) - \sum_{i=0}^{t-1} \eta(i)\left[\Pi_{j=i+1}^{t} (I - \eta(j)H)\right] \xi(i) - \eta(t)\xi(t) \]

    Taking expectation on both the sides, we get

    \[ \E[w(t+1)] = \left[\Pi_{i=0}^t (I - \eta(i)H)\right] w(0) \]

    This shows that $\E[w_{R_1}(t)] = \E[w_{R_2}(t)] = w_{GD}(t)$.

    Now, consider the loss given by $\mathcal{L}(w(t+1)) = w(t+1)^\top H w(t+1)$. Expanding it, we get three terms.

    The first one is the loss of the GD trajectory given by \[ \left(\left[\Pi_{i=0}^t (I - \eta(i)H)\right] w(0)\right)^\top H \left(\left[\Pi_{i=0}^t (I - \eta(i)H)\right] w(0)\right) \]

    The second term is the cross term between the noise and $w(0)$ term, given by

    \[ 2\left(-\sum_{i=0}^{t-1} \eta(i)\left[\Pi_{j=i+1}^{t} (I - \eta(j)H)\right] \xi(i) - \eta(t)\xi(t)\right)^\top H \left(\left[\Pi_{i=0}^t (I - \eta(i)H)\right] w(0)\right) \]

    Clearly the expectation of this term is zero as $\E[\xi(i)] = 0$ for all $i$.

    The third term is the cross term between the noise terms given by

    \[ \left(\sum_{i=0}^{t-1} \eta(i)\left[\Pi_{j=i+1}^{t} (I - \eta(j)H)\right] \xi(i) + \eta(t)\xi(t)\right)^\top H \left(\sum_{i=0}^{t-1} \eta(i)\left[\Pi_{j=i+1}^{t} (I - \eta(j)H)\right] \xi(i) + \eta(t)\xi(t)\right) \]

    Here, any term of the form $\xi(i)^\top M \xi(j)$ for some matrix $M$ where $i \neq j$ has $0$ expectation by independence and zero mean of $\xi(i)$ and $\xi(j)$. For the terms of the form $\xi(i)^\top M \xi(i)$, expanding $\xi(i)$ in the eigenbasis of $H$, we get that expectation of the third term scales with $\sigma^2$.

    Thus, we can see, for two different runs $R_1$ and $R_2$ satisfying $\sigma_1 \geq \sigma_2$, $\E[\mathcal{L}(w_{R_1}(t))] \geq \E[\mathcal{L}(w_{R_2}(t))]$.  
    
\end{proof}

\section{Additional Plots}

\subsection{Additional Plots for Section~\ref{sec:2}}\label{app:add-2}

\begin{figure}[H]
    \centering
    \includegraphics[width=0.45\linewidth]{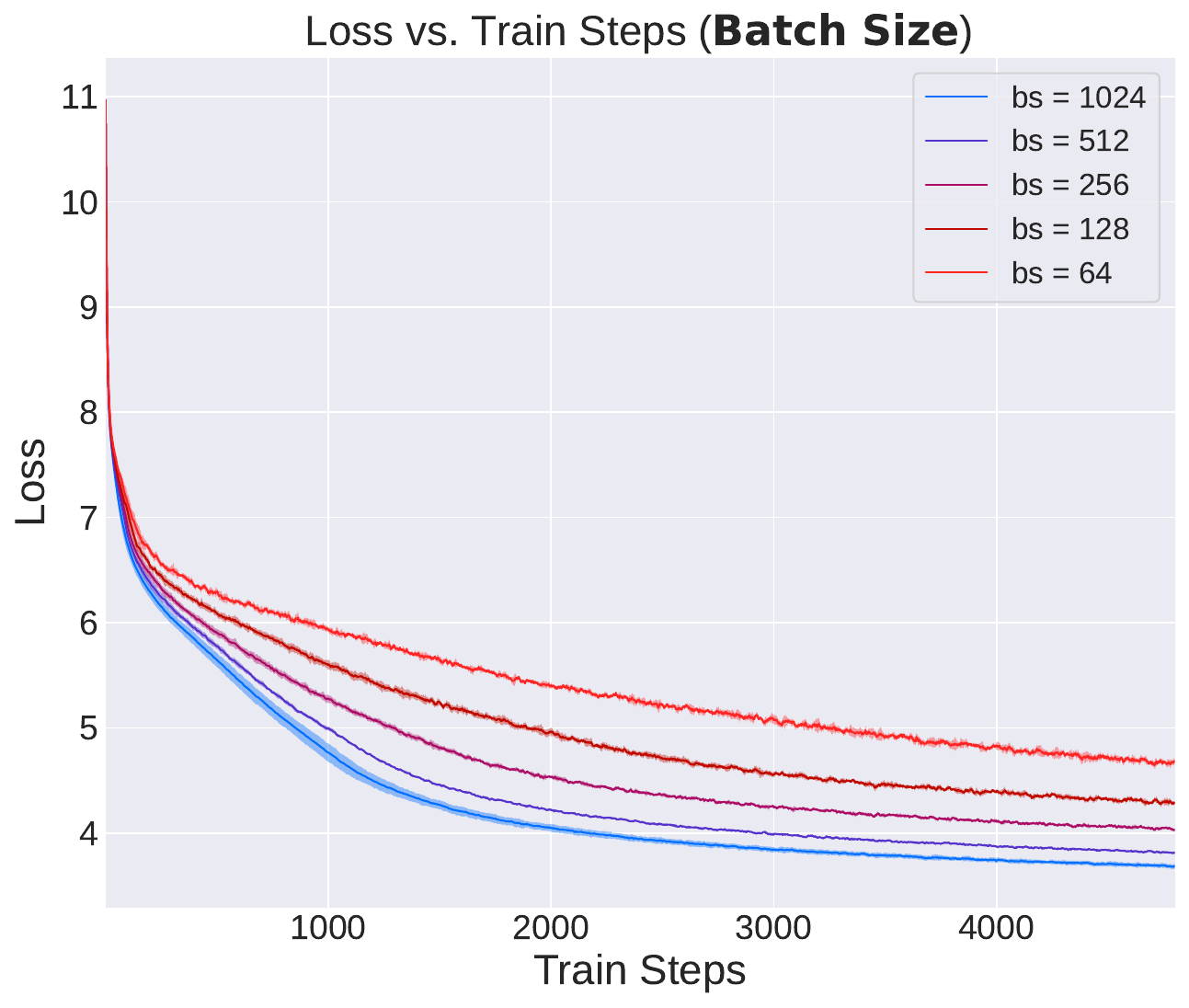}
    \caption{Experiments for online learning on the C4 dataset across several batch sizes with error bars. Consistent with our findings, runs with a lower amount of SGD noise yield better performance.}
    \label{fig:c4_multiple_seeds}
\end{figure}

\begin{figure}[H]
     \centering
         \includegraphics[width=0.6\linewidth]{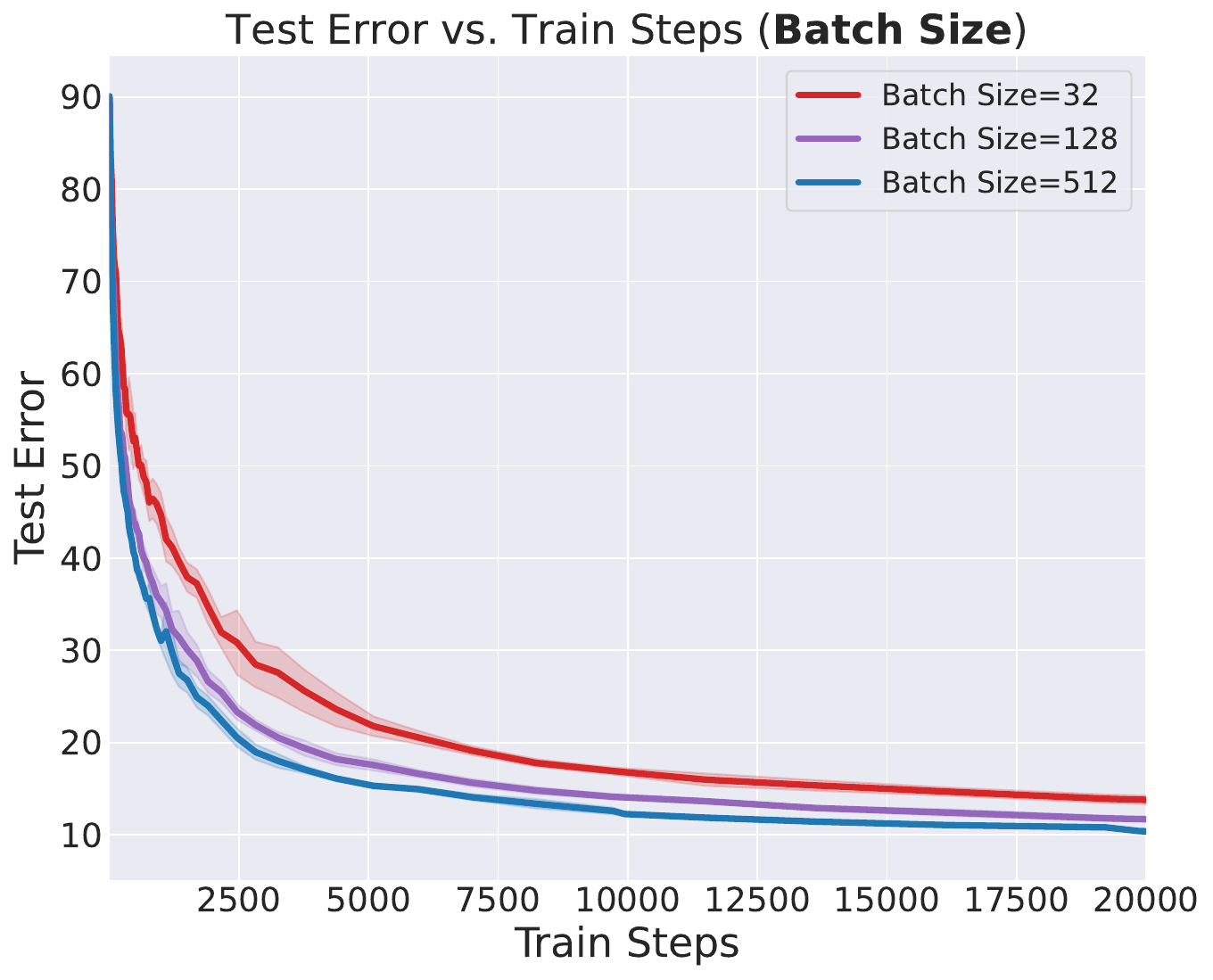}
         \caption{Test performance with error bars for CIFAR-5m in the online setting varying batch size. Note that all figures in Section~\ref{sec:2} were already averaged over $\geq 4$ runs.}
         \label{fig:cifar_multiple_seeds} 
\end{figure}

In Figures~\ref{fig:c4_multiple_seeds} and \ref{fig:cifar_multiple_seeds} we show a subset of figures from Section~\ref{sec:2} with error bars. Note that all figures in Section~\ref{sec:2} were already averaged over $\geq 4$ runs.

%In Figure~\ref{fig:7a_linear} we jointly decrease learning rate and increase batch size and show that the loss curves converge (with the x-axis as LR-normalized steps) while maintaining the observation that lower SGD noise runs perform better.

\begin{figure}[H]
         \centering
         \includegraphics[width=0.6\linewidth]{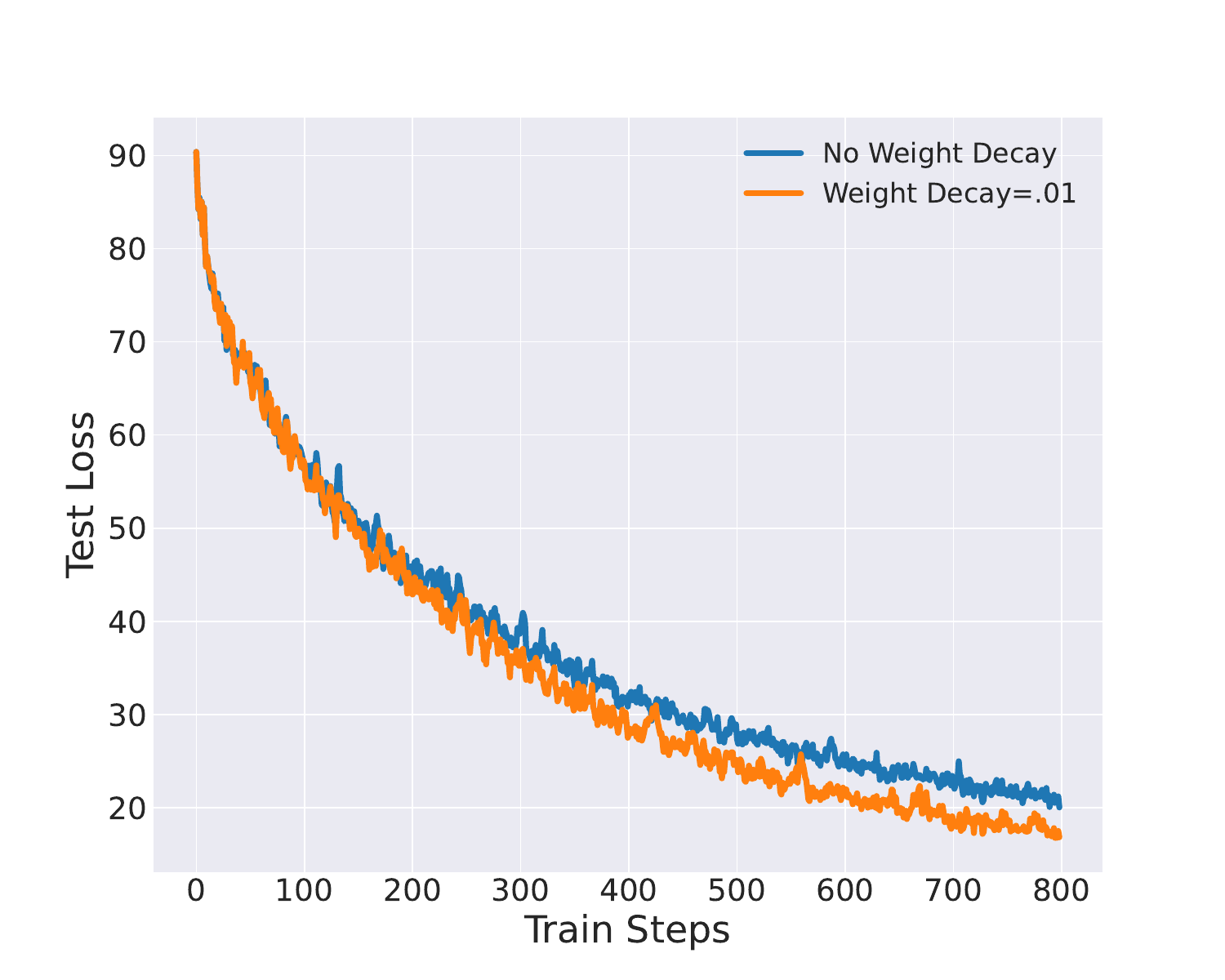}
         \caption{Explicit regularization can help in online learning. Specifically when training on the CIFAR 5m dataset, the test loss for a run using a small amount of weight decay is lower than that of a run with no weight decay.}
         \label{fig:prob1_6_1}
\end{figure}

In Figure~\ref{fig:prob1_6_1} we show an example in which explicit regularization helps in online learning. This is unlike the implicit regularization of SGD noise which does not help in our experiments in online learning.

% \begin{figure}[!ht]
%      \centering
%          \includegraphics[width=0.7\linewidth]{images/cifar-5m/gradientflow.pdf}
%          \caption{Convergence to gradient flow: we show that as learning rate decreases and batch size increases the loss curves converge (with x-axis as LR-normalized steps). Consistent with our findings, the lower SGD noise runs continue to perform better.}
%          \label{fig:7a_linear} 
% \end{figure}

\begin{figure}[H]
         \centering
         \includegraphics[width=0.6\linewidth]{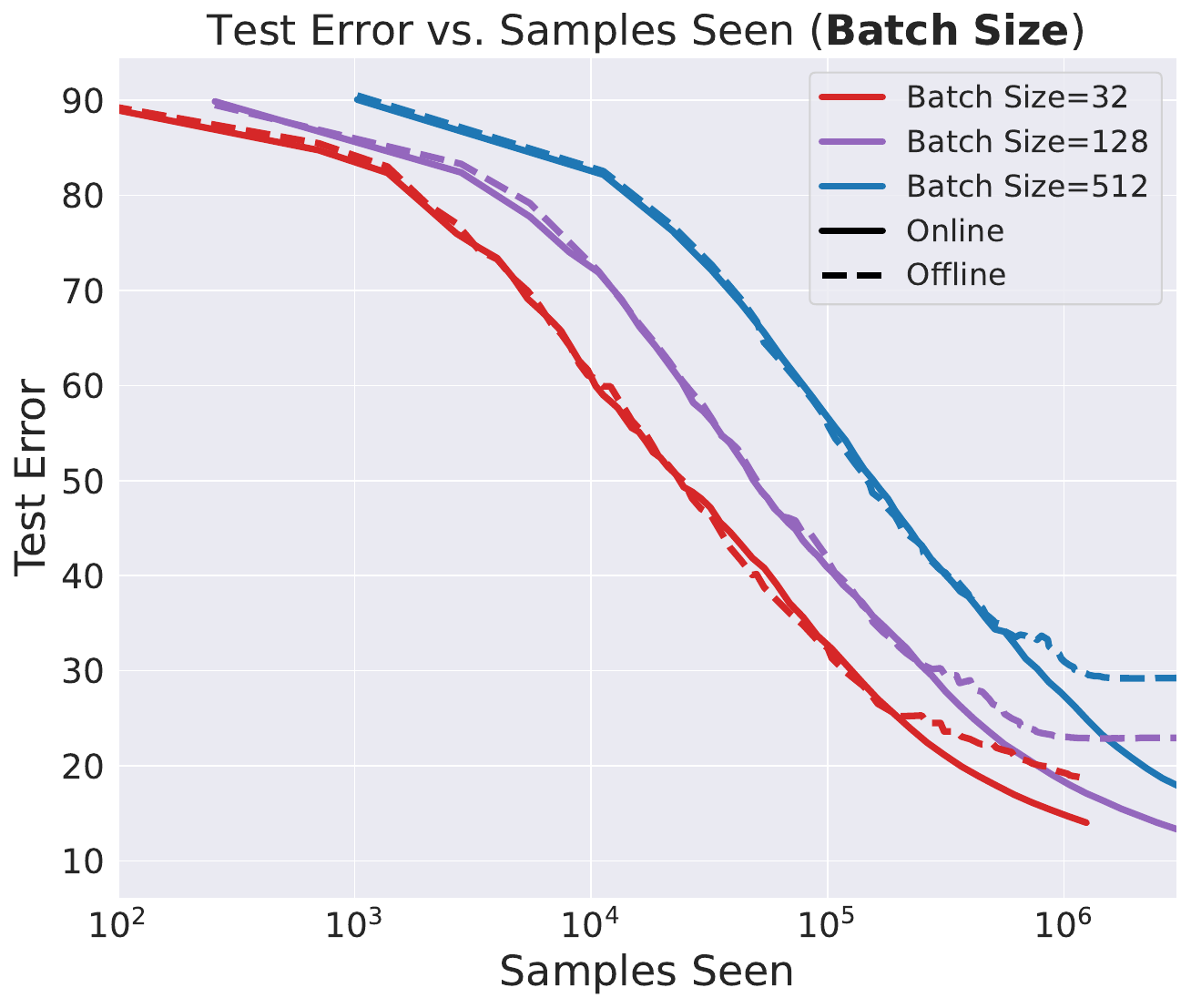}
         \caption{Test performance for Resnet18 on CIFAR-5m across varying batch sizes, with the number of examples seen in offline and online regime. The trend with number of examples remains similar in offline and online regime.}
         \label{fig:samples_seen}
\end{figure}

In Figure~\ref{fig:samples_seen} we plot Figure~\ref{fig:lr-c5m} but with x-axis as number of samples seen by the model rather than number of gradient steps. In this case we observe that larger batch sizes are worse.

\subsection{Additional Plots for Section~\ref{sec:snaploss}}\label{app:add-3}
\begin{figure}[H]
         \centering
         \includegraphics[width=0.6\linewidth]{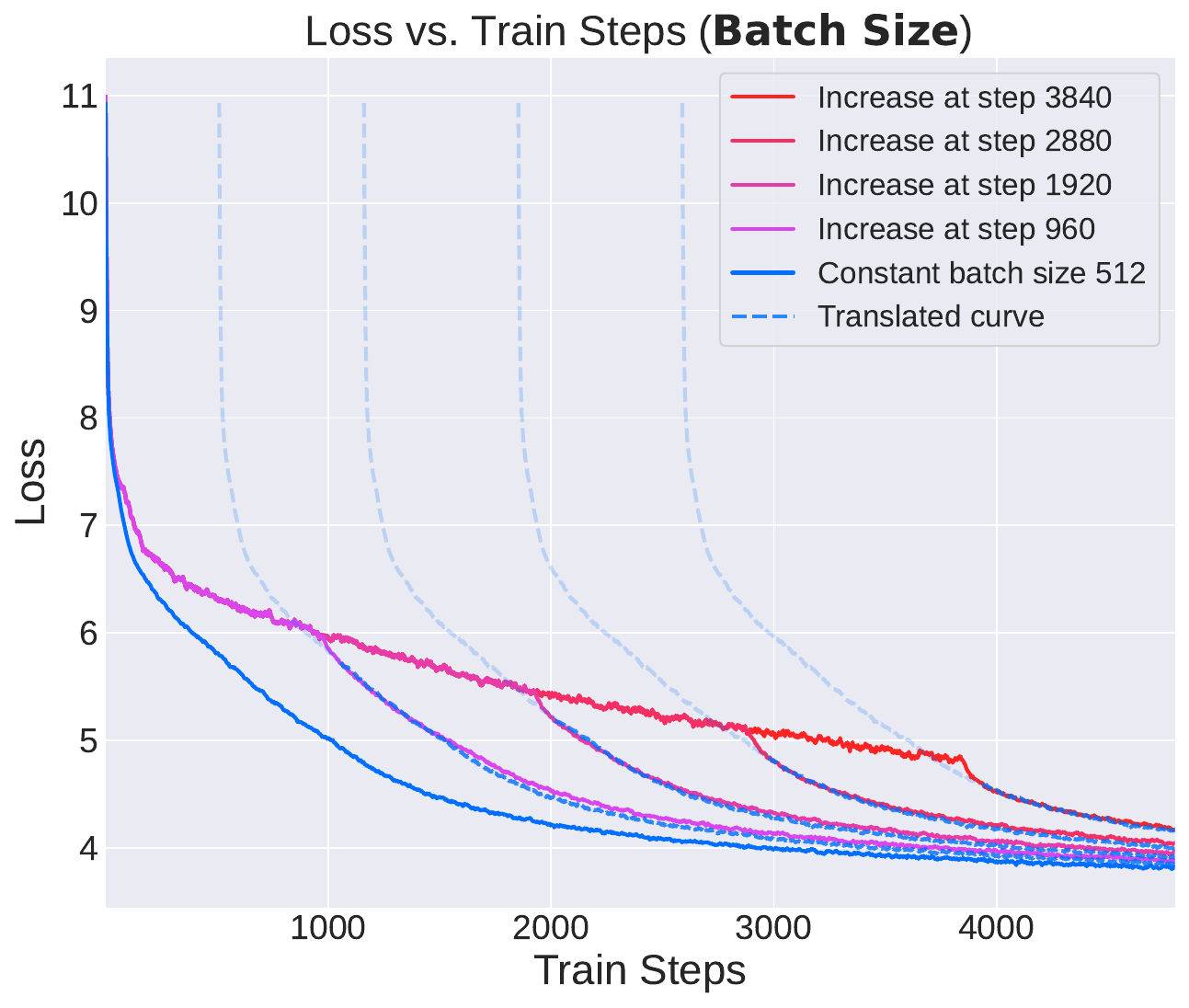}
         \caption{Experiments when switching the batch size from 64 to 512 at multiple points in training. As detailed in Section~\ref{sec:pointwise}, matching losses shortly after increasing the batch size results in a translated curve corresponding to the constant high batch size run.}
         \label{fig:bsz_switch_all}
\end{figure}

In Figure~\ref{fig:bsz_switch_all} we perform the same experiments in Section~\ref{sec:snaploss} on C4 but sweeping across the timestep when the batch size is increased from 64 to 512. We see that the loss curves shortly after increasing the batch size match a translated curve of the run with a constant batch size of 512.

\begin{figure}[H]
         \centering
         \includegraphics[width=0.6\linewidth]{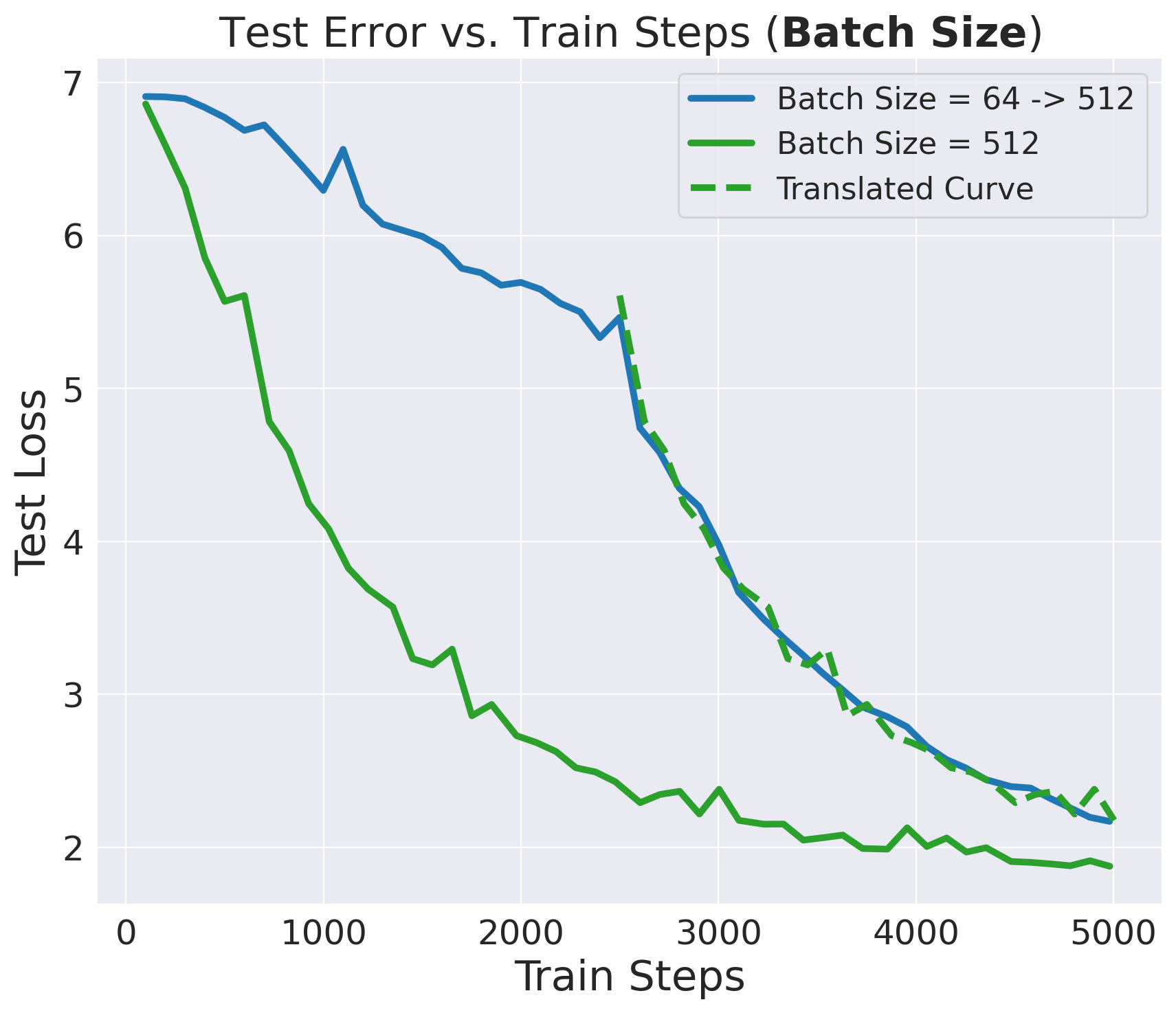}
         \caption{Changing SGD noise - increasing batch size during training for ConvNext on Imagenet. The red curves correspond to models trained with high SGD noise from initialization, and the blue curves trained with low SGD noise from initialization. In left plot the batch size is increased after $T_0$ steps while in right plot the batch size is decreased after $T_0$ steps. Across both experiments, changing batch size causes the original curve to follow a translated version (dashed) of new batch size curve}
         \label{fig:imagenet_bsz_change}
\end{figure}

In Figure~\ref{fig:imagenet_bsz_change} we do the batch size increasing experiment for ImageNet.

\subsection{Additional Plots for Section~\ref{sec:pointwise}}\label{app:add-4}
\begin{figure}[H]
         \centering
         \includegraphics[width=0.6\linewidth]{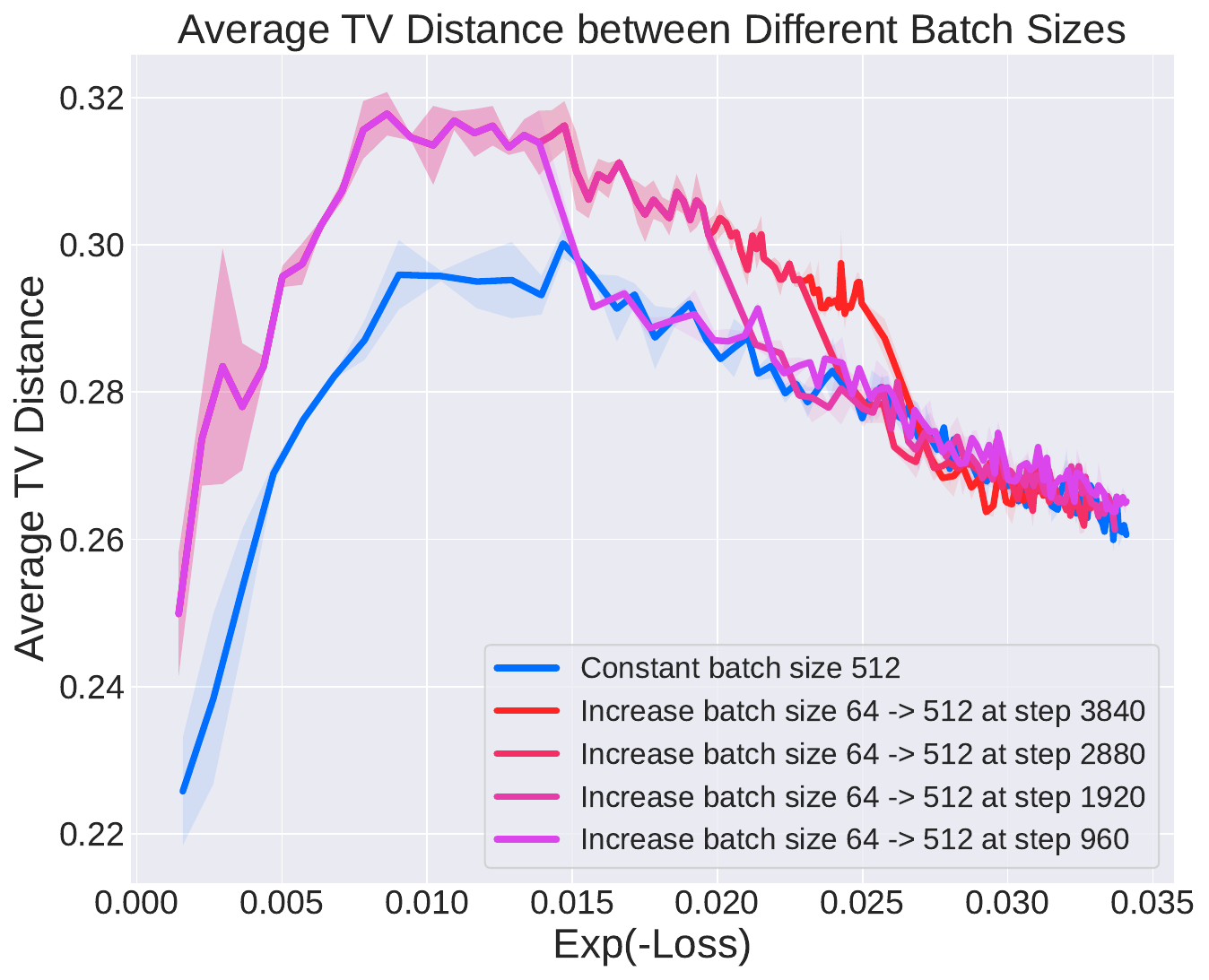}
         \caption{Examining the average TV distance between models when switching the batch size from 64 to 512 at multiple points in training. We can see that shortly after the increase in batch size, the average TV distance between this model and a model which has been trained with a constant batch size of 512 drops to match the average TV distance between two models trained with batch size 512 on two different seeds.}
         \label{fig:bsz_functional_all}
\end{figure}

As in Appendix~\ref{app:add-3}, we perform the same experiments from Section~\ref{sec:pointwise} across different timesteps for when the batch size is increased. The lines are averaged across three pairs of different seeds. As seen in Figure \ref{fig:bsz_functional_all}, the results shown in Section~\ref{sec:pointwise} are consistent across all different points in which we increase the batch size during training.

\begin{figure}[H]
         \centering
         \includegraphics[width=0.6\linewidth]{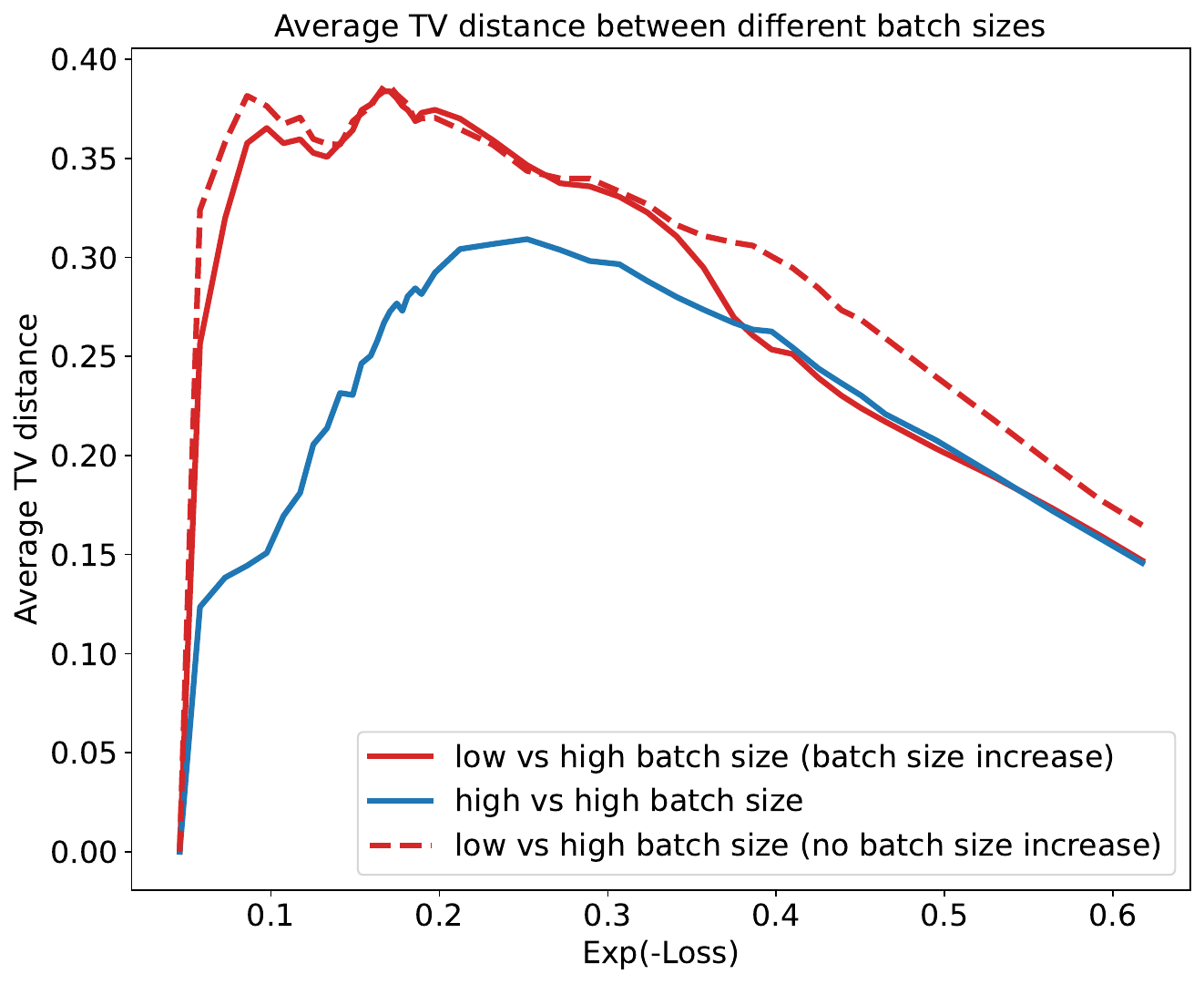}
         \caption{Average TV distance behavior on CIFAR-5m with and without reducing SGD noise, starting from the same initialization. For CIFAR-5m, we increase the batch size from 32 to 1024.  The dashed line represents the run with constant high SGD noise as a baseline. The figure demonstrates that the distance between the model with reduced noise and a low-noise model at the same loss is nearly identical to the distance between two random low-noise models trained with the same hyperparameters. This is in contrast with the distance between a high noise and a low-noise model at the same loss. This observation supports the ``golden path hypothesis'' and contradicts the ``fork in the road'' hypothesis. }
         \label{fig:bs_1024_same_init}
\end{figure}

\begin{figure}[H]
         \centering
         \includegraphics[width=0.6\linewidth]{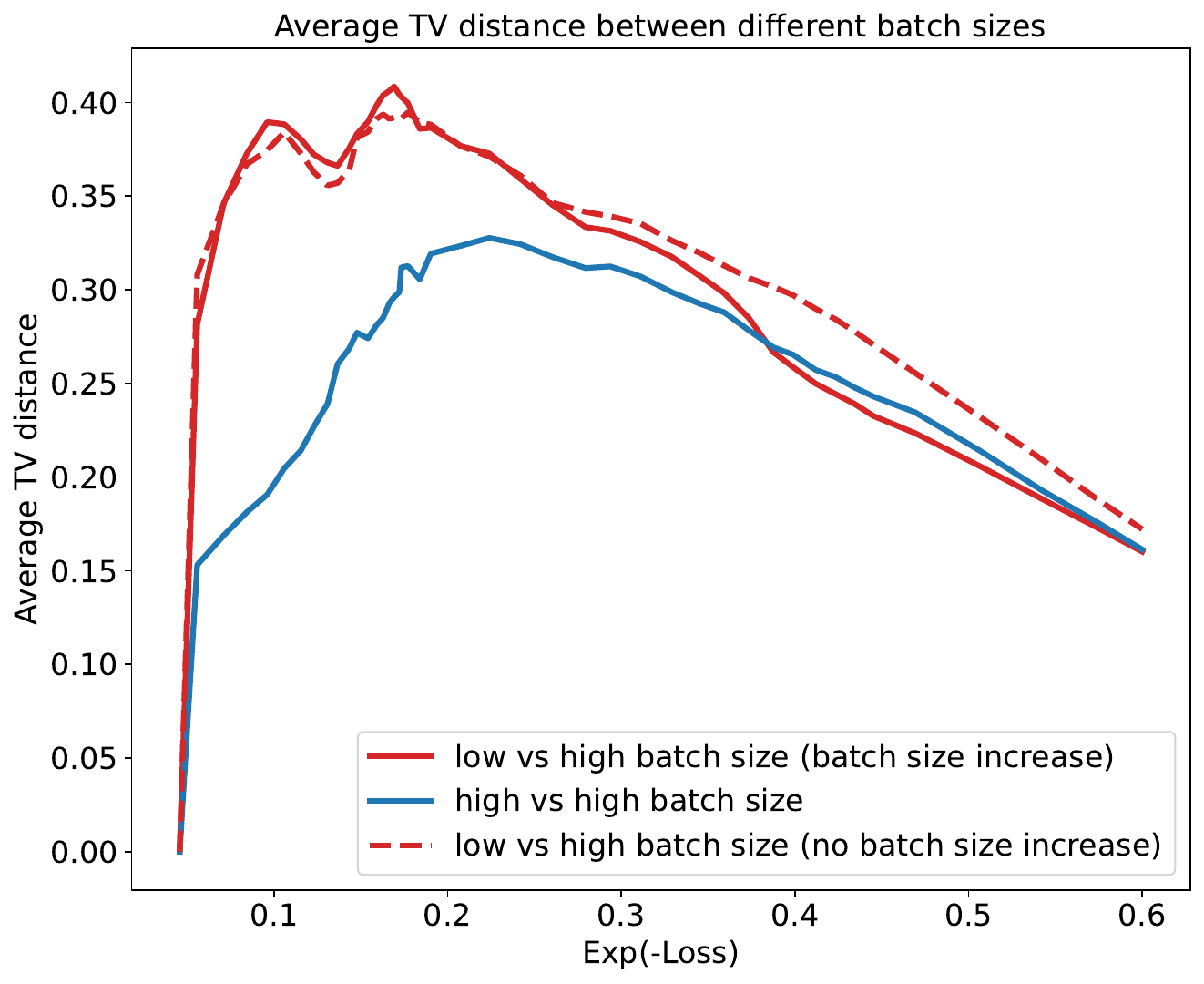}
         \caption{Average TV distance behavior on CIFAR-5m with and without reducing SGD noise, starting from the same initialization. For CIFAR-5m, we increase the batch size from 32 to 512.  The dashed line represents the run with constant high SGD noise as a baseline. The figure demonstrates that the distance between the model with reduced noise and a low-noise model at the same loss is nearly identical to the distance between two random low-noise models trained with the same hyperparameters. This is in contrast with the distance between a high noise and a low-noise model at the same loss. This observation supports the ``golden path hypothesis'' and contradicts the ``fork in the road'' hypothesis.}
         \label{fig:bs_512_same_init}
\end{figure}

Moreover, in Figure \ref{fig:bs_512_same_init} and \ref{fig:bs_1024_same_init}, we repeat the experiments of Section \ref{sec:pointwise} for CIFAR-5m, but with different runs starting from the same initialization. This removes the initialization noise, and purely takes SGD noise into account.

\end{document}